\documentclass[10pt,twocolumn,letterpaper]{article}

\usepackage[pagenumbers]{cvpr} 

\usepackage{graphicx}
\usepackage{amsmath}
\usepackage{amssymb}
\usepackage{booktabs}
\usepackage{amsthm}

\usepackage{algorithm, algorithmic}
\usepackage{multirow}
\usepackage{makecell}

\usepackage[pagebackref,breaklinks,colorlinks]{hyperref}

\usepackage[capitalize]{cleveref}
\crefname{section}{Sec.}{Secs.}
\Crefname{section}{Section}{Sections}
\Crefname{table}{Table}{Tables}
\crefname{table}{Tab.}{Tabs.}

\def\cvprPaperID{10757} 
\def\confName{CVPR}
\def\confYear{2023}

\begin{document}

\title{Diffusion Model for Generative Image Denoising}

\author{Yutong Xie \\
Peking University 
\and
Mingze Yuan \\
Peking University 
\and
Bin Dong \\
Peking University 
\and 
Quanzheng Li \\
Massachusetts General Hospital and Harvard Medical School
}
\maketitle

\begin{abstract}
In supervised learning for image denoising, usually the paired clean images and noisy images are collected or synthesised to train a denoising model. L2 norm loss or other distance functions are used as the objective function for training. It often leads to an over-smooth result with less image details. In this paper, we regard the denoising task as a problem of estimating the posterior distribution of clean images conditioned on noisy images. We apply the idea of diffusion model to realize generative image denoising. According to the noise model in denoising tasks, we redefine the diffusion process such that it is different from the original one. Hence, the sampling of the posterior distribution is a reverse process of dozens of steps from the noisy image. We consider three types of noise model, Gaussian, Gamma and Poisson noise. With the guarantee of theory, we derive a unified strategy for model training. Our method is verified through experiments on three types of noise models and achieves excellent performance.

\end{abstract}

\section{Introduction}
\label{sec:intro}

Image denoising \cite{dabov2006image, ramani2008monte, zhang2017beyond} has been studied for many years. Suppose $\boldsymbol{x}$ is a clean image and $\boldsymbol{y}$ is a noisy image. The noise model can be written as follows:
\begin{equation}
\label{eq:noise-model-definition}
    \boldsymbol{y} = \mathbf{N} \left( \boldsymbol{x}; \boldsymbol{z}, \text{params} \right),
\end{equation}
where $\boldsymbol{z}$ represents the source of noise, $\text{params}$ represents the parameters of the noise model. Current supervised learning methods focus on training a denoising model using paired clean images and noisy images. Hence, collecting or synthesising such pairs as training data is important. Usually, the denoising model is trained by the following loss function:
\begin{equation}
\label{eq:sl-training}
L(\theta) = \mathbb{E}_{\boldsymbol{x}, \boldsymbol{y}} \left[ \mathrm{d} \left(\boldsymbol{x}, f(\boldsymbol{y}; \theta) \right) \right],
\end{equation}
where $f(\cdot; \theta)$ is a neural network and $\mathrm{d} \left( \cdot, \cdot\right)$ is a distance metric. This methodology of supervised learning in essence is to define the denoising task as a training problem of determined mapping, from the noisy image $\boldsymbol{y}$ to the clean one $\boldsymbol{x}$. However, the denoising model trained in this manner often leads to a result with average effect. From the Bayesian perspective, $\boldsymbol{y}$ conditioned on $\boldsymbol{x}$ follows a posterior distribution:
\begin{equation}
\label{eq:bayesian}
q \left(\boldsymbol{x} \mid \boldsymbol{y} \right) = \frac{q \left( \boldsymbol{y} \mid \boldsymbol{x} \right) q \left(\boldsymbol{x} \right)}{q \left( \boldsymbol{y}\right)}.
\end{equation}
When $\mathrm{d}(\cdot, \cdot)$ in \cref{eq:sl-training} is L2 norm distance, the trained model will be an estimation of $\mathbb{E} \left[ \boldsymbol{x} \mid \boldsymbol{y}\right]$, \ie the posterior mean. This explains why denoised result in usual supervised learning is over-smooth.

\begin{figure}[t]
  \centering
    \includegraphics[width=\linewidth]{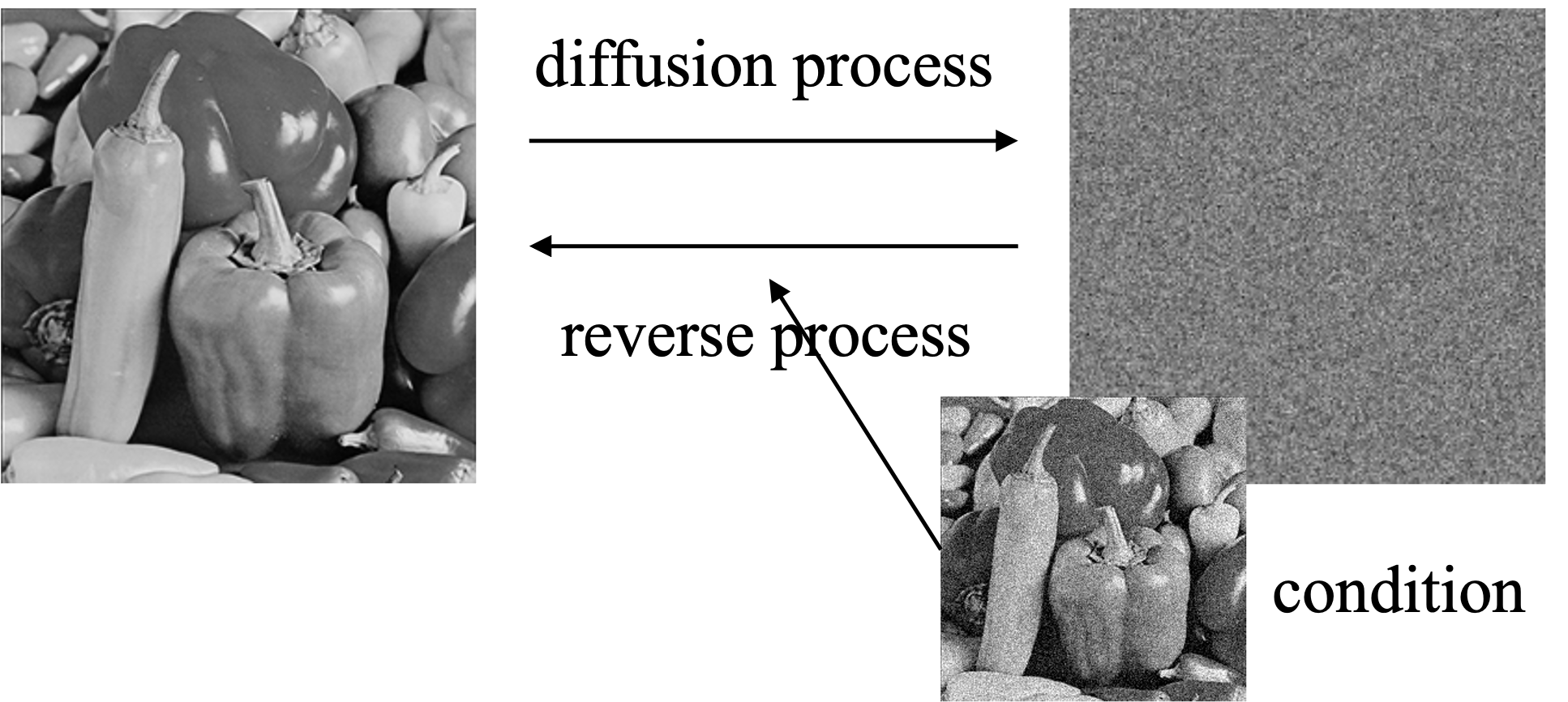}
    \caption{The illustration of applying the original diffusion model to image denoising.}
    \label{fig:demo-1}
\end{figure}

\begin{figure}[t]
  \centering
    \includegraphics[width=\linewidth]{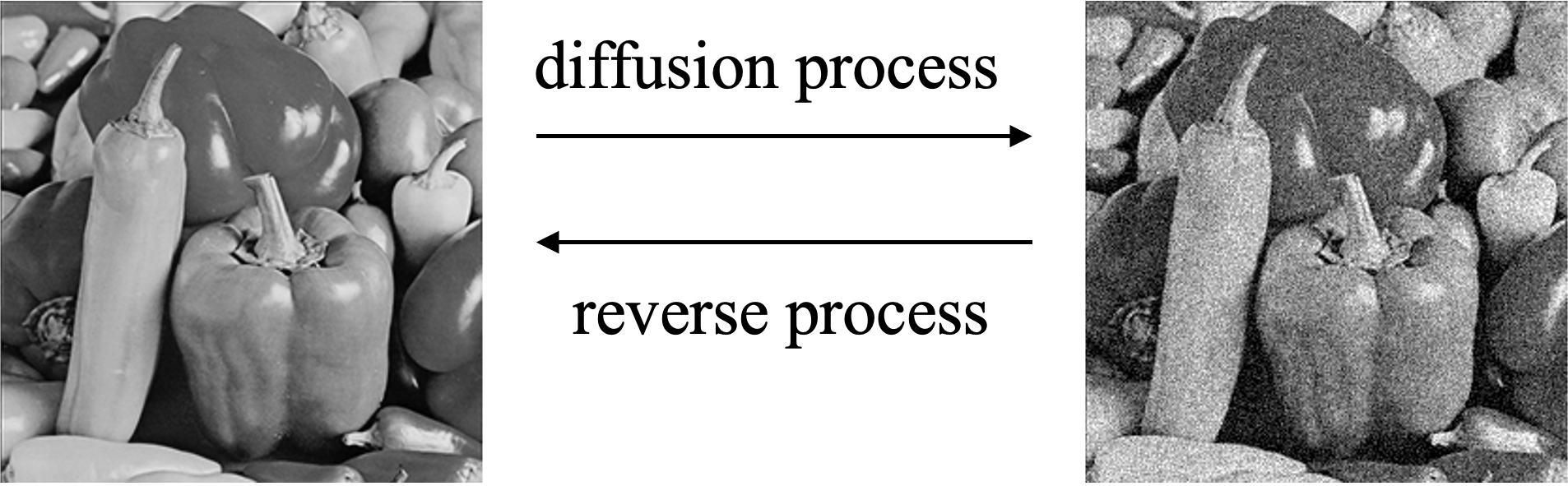}
    \caption{The illustration of proposed diffusion model for generative image denoising.}
    \label{fig:demo-2}
\end{figure}

To avoid the average effect, we regard image denoising as a problem of estimation of posterior distribution $q \left( \boldsymbol{x} \mid \boldsymbol{y}\right)$. Hence, we do not train a denoising model representing a determined mapping. Instead, we train a generative denoising model. Recently, the diffusion model has achieve tremendous success in the domain of image generative tasks \cite{sohl2015deep, ho2020denoising, song2020score, kong2020diffwave, saharia2021image, meng2021sdedit}. In the original diffusion model, diffusion process transform clean images $\boldsymbol{x}$ to total Gaussian noise by adding little Gaussian noise and reducing the signal of $\boldsymbol{x}$ step by step. The sampling of target distribution is realized by a reverse process with hundreds and thousands of iterations from total Gaussian noise to clean images. Though it is a powerful generative model, applying the diffusion model directly to image denoising is not a desirable way. Its illustration is shown in \cref{fig:demo-1}. It is time-consuming without any acceleration trick and does not fully utilized the residual information in noisy images. Interestingly, we observe that the diffusion process is similar to the noise model defined in \cref{eq:noise-model-definition}, though the noise in diffusion model is limited to Gaussian noise and the signal of $\boldsymbol{x}$ is reduced along the diffusion process. If the diffusion process is completely consistent with the noise model, it is possible that we can begin the reverse process with noisy images, rather than total Gaussian noise and reduce the iteration greatly. For this purpose, we propose a new diffusion model for denoising tasks. The diffusion process in our method is designed according to the specific noise model such that they are consistent. As a result, the reverse process can start from the noisy image to realize sampling of $q \left(\boldsymbol{x} \mid \boldsymbol{y} \right)$. We show the idea in \cref{fig:demo-2}. The details of design for diffusion process and reverse process, model training and sampling algorithms are described in \cref{sec:method}.

In summary, our main contributions are: (1) We propose a new diffusion model designed for image denoising tasks. (2) We design the diffusion process, model training strategy and sampling algorithms for three types noise models, Gaussian, Gamma and Poisson. (3) Our experiments show that our proposed method is feasible.


\section{Related Works}
\label{sec:related_works}

\subsection{Supervised Learning}

Learning from paired noisy-clean data is the mainstream in image denoising. Given paired noisy-clean data, it is straightforward to train supervised denoising methods. Albeit breakthrough performance has been achieved, the success of most existing deep denoisers heavily depend on supervised learning with large amount of paired noisy-clean images, which can be further divided into synthetic pairs \cite{zhang2017beyond, zhang2018ffdnet, zhou2020awgn, yue2019variational, guo2019toward, zamir2020cycleisp, kim2020transfer, li2021cross} and realistic pairs \cite{anwar2019real, cheng2021nbnet, hu2021pseudo, liu2021invertible, ren2021adaptive, zamir2020learning, zheng2021deep}.

\subsection{Original Diffusion Model}

In the original diffusion model \cite{ho2020denoising}, the diffusion process is defined as:
\begin{equation}
\boldsymbol{x}_{t+1} = \alpha_{t} \boldsymbol{x}_{t} + \beta_{t} \boldsymbol{z}, t = 0, ..., 
\end{equation}
where $\boldsymbol{x}_{0}$ is the clean image and $\boldsymbol{z}$ follows the standard multi-variable Gaussian distribution. $\alpha_{t}$ is smaller than but near $1$, and $\beta_{t}$ is a very small value. When $T$ is large enough, $\boldsymbol{x}_{T}$ approximately follows the standard multi-variable Gaussian distribution. The reverse process is the inverse of the diffusion process. Therefore, the reverse process from a random Gaussian noise will lead to a sample of $q \left( \boldsymbol{x}_{0}\right)$.

\section{Method}
\label{sec:method}

In this section, we provide a more detailed description of the presented approach. The organization is as follows: In \cref{sec:framework} we introduce the basic framework of diffusion model designed for image denoising tasks. We present the application for three kinds of noise models in \cref{sec:application}. Finally, \cref{sec:discussion} is the further discussion. All full proofs and derivation in this section can be found in Appendix.

\subsection{Basic Framework}
\label{sec:framework}

Suppose the noise model is known and we present it using following form:
\begin{equation}
\label{eq:noise-model}
    \boldsymbol{y} = \mathbf{N} \left( \boldsymbol{x}; \boldsymbol{z}, \text{params} \right),
\end{equation}
where $\boldsymbol{z}$ represents the source of noise, $\text{params}$ represents the parameters of the noise model. Here, $\boldsymbol{y} \sim \mathbf{y}$ is the noisy image and $\boldsymbol{x} \sim \mathbf{x}$ is the clean image. In this paper, we use different font to distinguish random variables and their samples like $\mathbf{x}$ and $\boldsymbol{x}$. Our target is to realize sampling of $q \left( \boldsymbol{x} \mid \boldsymbol{y} \right)$.

Since we adapt the idea of the original diffusion model, next we introduce the definition of diffusion process and reverse process.

\subsubsection{Diffusion Process}

Let $t = 0, 1, ..., N$, we construct $N+1$ random variables, $\mathbf{x}_{t}$, through a sequence of parameters in \cref{eq:noise-model}, $\left\{ \text{params}_{1}, \text{params}_{2}, ...,  \text{params}_{N} \right\} $. Here, we define that $\mathbf{x}_{0} = \mathbf{x}$, and let $\text{params}_{N}$ be $\text{params}$ in \cref{eq:noise-model}. Given $\boldsymbol{x}_{0} \sim \mathbf{x}_{0}$, we have the following definition for diffusion process along $t$:
\begin{equation}
\label{eq:definition-diffusion}
\boldsymbol{x}_{t} \mid \mathbf{x}_{0} = \mathbf{N} \left( \boldsymbol{x}_{0}; \boldsymbol{z}, \text{params}_{t} \right), t = 1, ..., N.
\end{equation}
Such definition indicates that $ \mathbf{x}_{N} = \mathbf{y}$. In the rest of this paler, for the sake of convenience we use $\mathbf{x}_{0}$ and $\mathbf{x}_{N}$ to represent $\mathbf{x}$ and $\mathbf{y}$ respectively. Thus, $\boldsymbol{x}_{0}$ is a clean image and $\boldsymbol{x}_N$ is the noisy image to be denoised. Usually, the sequence of $\left\{ \text{params}_{t}\right\} $ can be regarded as a discrete sampling of a continuous (and monotonous) function $\text{Params}(t)$.

According to \cref{eq:definition-diffusion}, $\boldsymbol{x}_{t}, t = 1, ..., N-1$ are also noisy images with a noise level different from $\boldsymbol{x}_{N}$. When $\text{params}_{t}, t = 1, ..., N$, is fixed, the distribution of $\mathbf{x}_{t}$ has been determined by the distribution of $\mathbf{x}_{0}$ and the noise model. We have
\begin{equation}
q \left(\boldsymbol{x}_{t} \right) = \int_{\mathbf{x}_{0}} q \left( \boldsymbol{x}_{t} \mid \boldsymbol{x}_{0}\right) q \left(\boldsymbol{x}_{0} \right)  \mathrm{d} \boldsymbol{x}_{0},
\end{equation}
where $q \left(\boldsymbol{x}_{0} \right)$ is the probability density function of $\mathbf{x}_{0}$ and $q \left( \boldsymbol{x}_{t} \mid \boldsymbol{x}_{0}\right)$ is related to the noise model defined in \cref{eq:definition-diffusion}. However, the relation between $\mathbf{x}_{t}, i = 1, ..., N$ is not defined. Here, we do not provide the specific assumption for the relation and we discuss it in \cref{sec:application}. Nevertheless, the diffusion process is related to the noise model in denoising task by the definition above.

\subsubsection{Reverse Process}

Since our target distribution is $q \left( \boldsymbol{x}_{0} \mid \boldsymbol{x}_{N} \right)$ where $\boldsymbol{x}_{N}$ is the given noisy image. Suppose the model is represented by $p_{\theta} \left( \boldsymbol{x}_{0} \mid \boldsymbol{x}_{N} \right)$. 
We define the reverse process as a Markov chain:
\begin{equation}
\label{eq:reverse-markov}
    p_{\theta}\left(\boldsymbol{x}_{t} \mid \boldsymbol{x}_{t+1}, ..., \boldsymbol{x}_{N} \right) = p_{\theta}\left(\boldsymbol{x}_{t} \mid \boldsymbol{x}_{t+1}, \boldsymbol{x}_{N} \right).
\end{equation}
Then, we have the following derivation:
\begin{equation}
\label{eq:reverse-sampling}
\begin{split}
& p_{\theta} \left( \boldsymbol{x}_{0} \mid \boldsymbol{x}_{N} \right) \\
= & \int_{\mathbf{x}_{1:N-1}} p_{\theta} \left( \boldsymbol{x}_{0}, \boldsymbol{x}_{1}, ..., \boldsymbol{x}_{N-1} \mid \boldsymbol{x}_{N} \right) \mathrm{d} \boldsymbol{x}_{1} \cdots \mathrm{d} \boldsymbol{x}_{N-1} \\
= & \int_{\mathbf{x}_{1:N-1}} \prod_{t=0}^{t=N-1} p_{\theta} \left( \boldsymbol{x}_{t} \mid \boldsymbol{x}_{t+1}, ..., \boldsymbol{x}_{N} \right) \mathrm{d} \boldsymbol{x}_{1} \cdots \mathrm{d} \boldsymbol{x}_{N-1} \\
= & \int_{\mathbf{x}_{1:N-1}} \prod_{t=0}^{t=N-1} p_{\theta} \left( \boldsymbol{x}_{t} \mid \boldsymbol{x}_{t+1}, \mathbf{x}_{N} \right) \mathrm{d} \boldsymbol{x}_{1} \cdots \mathrm{d} \boldsymbol{x}_{N-1}.
\end{split}
\end{equation}
Here, \cref{eq:reverse-markov} is utilized in the third equation. The subscript $1:N-1$ in the notation $\mathbf{x}_{1:N-1}$ is an abbreviation for $\left\{\mathbf{x}_{1}, ..., \mathbf{x}_{N-1} \right\}$. For the sake of convenience, we continue to use this abbreviation in the rest of this paper. \Cref{eq:reverse-sampling} indicates that we realize the sampling of $q \left( \boldsymbol{x}_{0} \mid \boldsymbol{x}_{N} \right)$ through iterative sampling of $p_{\theta} \left( \mathbf{x}_{t} \mid \mathbf{x}_{t+1}, \mathbf{x}_{N} \right)$. Therefore, we have the following sampling algorithm.

\begin{algorithm}
\renewcommand{\algorithmicrequire}{\textbf{Input:}}
\renewcommand{\algorithmicensure}{\textbf{Output:}}
\caption{The general sampling process}
\label{alg:general-sampling}
\begin{algorithmic}[1]
    \REQUIRE noisy image $\boldsymbol{x}_{N}$ and model $p_{\theta}$.
    \ENSURE ${\boldsymbol{x}}_{0}$.
    \FOR{$t = N-1, ..., 0$}
        \STATE Sample $\boldsymbol{x}_t$ by  $p_{\theta}\left(\boldsymbol{x}_{t} \mid \boldsymbol{x}_{t+1}, \boldsymbol{x}_{N} \right)$.
    \ENDFOR
\end{algorithmic}  
\end{algorithm}

\subsubsection{Derivation of Model Training}

In fact, $p_{\theta} \left( \boldsymbol{x}_{0} \mid \boldsymbol{x}_{N}\right)$ is a multiple--hidden-variable model with $\boldsymbol{x}_{t}, t = 1, ..., N-1$. Based on the definition of the diffusion process and reverse process, we can derive the evidence lower bound objective (ELBO) as follows:
\begin{equation}
\label{eq:general-loss}
\begin{split}
L &= \mathbb{E}_{\mathbf{x}_{0}, \mathbf{x}_{N}} \left[ - \log p_{\theta}\left( \mathbf{x}_{0} \mid \mathbf{x}_{N} \right) \right] \\
& \leq \mathbb{E}_{q_{0:N}}\left[- \log \frac{p_{\theta}\left( \mathbf{x}_{0:N-1} \mid \mathbf{x}_{N} \right) }{q \left( \mathbf{x}_{1:N-1} \mid \mathbf{x}_{0}, \mathbf{x}_{N} \right)}\right] \\
& = \mathbb{E}_{q_{0:N}}\left[- \log \frac{\prod_{t=0}^{t=N-1} p_{\theta} \left( \mathbf{x}_{t} \mid \mathbf{x}_{t+1}, \mathbf{x}_{N} \right)  }{q \left( \mathbf{x}_{1:N-1} \mid \mathbf{x}_{0}, \mathbf{x}_{N} \right)}\right].
\end{split}
\end{equation}
The proof is in Appendix. The further derivation of \cref{eq:general-loss} depends on the assumption of diffusion process and will be shown in \cref{sec:application}.

\subsection{Application}
\label{sec:application}

In this part, we discuss three types of noise models for image denoising, Gaussian, Gamma and Poisson noise, as examples. We will give the specific assumption for the diffusion process, derive the objective loss function for training, and show the full sampling algorithms.

\subsubsection{Gaussian Noise}
\label{sec:gaussian}

The Gaussian noise model in denoising task is defined by the following form:
\begin{equation}
    \boldsymbol{x}_{N} = \boldsymbol{}{x}_{0} + \sigma \boldsymbol{z}, \quad \boldsymbol{z} \sim \mathcal{N} \left( \boldsymbol{0}, \boldsymbol{I} \right),
\end{equation}
where $\mathcal{N} \left( \boldsymbol{0}, \boldsymbol{I} \right)$ is standard Gaussian distribution with independent components.

We select $0 = \sigma_{0} < \sigma_{1} < \sigma_{2} < \cdots < \sigma_{N-1} < \sigma_{N} = \sigma$. Therefore, $\left\{ \sigma_t \right\}$ is a monotonically increasing sequence. Let
\begin{equation}
\label{eq:gaussian-assumption}
\boldsymbol{x}_{t+1} = \boldsymbol{x}_{t} + \sqrt{\sigma_{t+1}^2 - \sigma_{t}^{2}} \boldsymbol{z}_{t+1}, \quad \boldsymbol{z}_{t+1} \sim \mathcal{N} \left( \boldsymbol{0}, \boldsymbol{I} \right),
\end{equation}
where $t = 0, 1, ..., N-1$. Then we have:
\begin{equation}
\begin{split}
\boldsymbol{x}_{t} &= \boldsymbol{x}_{0} + \left( \sum_{i=0}^{t-1} \sqrt{\sigma_{i+1}^{2} - \sigma_{i}^{2}} \boldsymbol{z}_{i} \right) \\
&= \boldsymbol{x}_{0} + \sigma_{t} \mathbf{z}, \quad \boldsymbol{z} \sim \mathcal{N} \left( \boldsymbol{0}, \boldsymbol{I} \right),
\end{split}
\end{equation}
where $t = 1, ..., N$. \Cref{eq:gaussian-assumption} defines the relation between $\mathbf{x}_{t}$, $t = 1, ..., N$ and we have
\begin{equation}
\label{eq:assumption-1}
q \left( \boldsymbol{x}_{t} \mid \boldsymbol{x}_{0:t-1} \right) = q \left( \boldsymbol{x}_{t} \mid \boldsymbol{x}_{t-1} \right).
\end{equation}
Thus, $\mathbf{x}_{t}$, $t = 1, ..., N$ are a Markov chain. By now, we define a full diffusion process for Gaussian Noise.

From \cref{eq:assumption-1}, we can derive the following two properties:
\begin{equation}
\label{eq:property-1-assumption-1}
\begin{split}
    q \left( \boldsymbol{x}_{1:N-1} \mid \boldsymbol{x}_{0}, \boldsymbol{x}_{N} \right)
    = \prod_{t=1}^{N-1} q \left( \boldsymbol{x}_{t} \mid \boldsymbol{x}_{0}, \boldsymbol{x}_{t+1} \right).
\end{split}
\end{equation}
\begin{equation}
\label{eq:property-2-assumption-1}
\begin{split}
    q \left( \boldsymbol{x}_{t} \mid \boldsymbol{x}_{t+1:N}\right)
    = q \left( \boldsymbol{x}_{t} \mid \boldsymbol{x}_{t+1} \right).
\end{split}
\end{equation}
The proof of \cref{eq:property-1-assumption-1} and \cref{eq:property-2-assumption-1} are in Appendix.

Applying \cref{eq:property-1-assumption-1} to \cref{eq:general-loss}, we have that:
\begin{equation}
\label{eq:obj-assumption-1}
\begin{split}
    L & \leq \mathbb{E}_{q_{0:N}}\left[- \log \frac{\prod_{t=0}^{t=N-1} p_{\theta} \left( \boldsymbol{x}_{t} \mid \boldsymbol{x}_{t+1}, \boldsymbol{x}_{N} \right)  }{\prod_{t=1}^{N-1} q \left( \boldsymbol{x}_{t} \mid \boldsymbol{x}_{0}, \boldsymbol{x}_{t+1} \right)}\right] \\
    & = \mathbb{E}_{q_{0:N}}\left[- \sum_{t=1}^{N-1} \log \frac{p_{\theta} \left( \boldsymbol{x}_{t} \mid \boldsymbol{x}_{t+1}, \boldsymbol{x}_{N} \right)  }{q \left( \boldsymbol{x}_{t} \mid \boldsymbol{x}_{0}, \boldsymbol{x}_{t+1} \right)} - \log p_{\theta} \left(\boldsymbol{x}_{0} \mid \boldsymbol{x}_{1}, \boldsymbol{x}_{N} \right) \right] \\
    & = \sum_{t=1}^{N-1} \mathbb{E}_{q_{0,t+1, N}} \left[ D_{\mathrm{KL}} \left( q \left( \boldsymbol{x}_{t} \mid \boldsymbol{x}_{0}, \boldsymbol{x}_{t+1} \right) \| p_{\theta} \left( \boldsymbol{x}_{t} \mid \boldsymbol{x}_{t+1}, \mathbf{x}_{N} \right)  \right) \right] \\
    & \quad \quad +  \mathbb{E}_{q_{0,1,N}} \left[  - \log p_{\theta} \left(\boldsymbol{x}_{0} \mid \boldsymbol{x}_{1}, \boldsymbol{x}_{N} \right) \right].
\end{split}
\end{equation}
Because \cref{eq:property-2-assumption-1} indicates that given $\boldsymbol{x}_{t+1}$, $\boldsymbol{x}_{t}$ is not dependent on $\boldsymbol{x}_{N}$ when $t < N - 1$. Therefore, we can further assume that
\begin{equation}
    p_{\theta} \left( \boldsymbol{x}_{t} \mid \boldsymbol{x}_{t+1}, \boldsymbol{x}_{N} \right) = p_{\theta} \left( \boldsymbol{x}_{t} \mid \boldsymbol{x}_{t+1} \right).
\end{equation}
As a result, \cref{eq:obj-assumption-1} is simplified as
\begin{equation}
\label{eq:simplified-obj-assumption-1}\
\begin{split}
    L & \leq \sum_{t=1}^{N-1} \mathbb{E}_{q_{0,t+1, N}} \left[ D_{\mathrm{KL}} \left( q \left( \boldsymbol{x}_{t} \mid \boldsymbol{x}_{0}, \boldsymbol{x}_{t+1} \right) \| p_{\theta} \left( \boldsymbol{x}_{t} \mid \boldsymbol{x}_{t+1} \right)  \right) \right] \\
    & \quad \quad + \mathbb{E}_{q_{0,1,N}} \left[  - \log p_{\theta} \left(\boldsymbol{x}_{0} \mid \boldsymbol{x}_{1} \right) \right].
\end{split}
\end{equation}

Now, we consider the main part of the loss function $\mathbb{E}_{q_{0,t+1, N}} \left[ D_{\mathrm{KL}} \left( q \left( \boldsymbol{x}_{t} \mid \boldsymbol{x}_{0}, \boldsymbol{x}_{t+1} \right) \| p_{\theta} \left( \boldsymbol{x}_{t} \mid \boldsymbol{x}_{t+1} \right)  \right) \right]$. We begin with the analytical form of $q \left( \boldsymbol{x}_{t} \mid \boldsymbol{x}_{0}, \boldsymbol{x}_{t+1} \right)$. We have the conclusion that:
\begin{equation}
\label{eq:gaussian-q-posterior}
\begin{split}
    q \left( \boldsymbol{x}_{t} \mid \boldsymbol{x}_{0}, \boldsymbol{x}_{t+1} \right) \sim \mathcal{N} \left( \tilde{\boldsymbol{\mu}}_{t}, \tilde{\sigma}_{t} \boldsymbol{I}  \right), t = 0, 1, ..., N-1,
\end{split}
\end{equation}
where 
\begin{equation}
\label{eq:gaussian-q-posterior-param}
\begin{split}
    \tilde{\boldsymbol{\mu}}_{t} &= \frac{\sigma_{t}^{2}}{\sigma_{t+1}^{2}} \boldsymbol{x}_{t+1} + \frac{\sigma_{t+1}^{2} - \sigma_{t}^{2}}{\sigma_{t+1}^{2}} \boldsymbol{x}_{0}, \\
    \tilde{\sigma}_{t}&= \frac{\sigma_{t}}{\sigma_{t+1}} \sqrt{\sigma_{t+1}^{2} - \sigma_{t}^{2}}.
\end{split}
\end{equation}
Therefore, we assume that $p_{\theta} \left( \boldsymbol{x}_{t} \mid \boldsymbol{x}_{t+1} \right)$, $t = 1, ..., N-1$, also follows a Gaussian distribution:
\begin{equation}
\label{eq:gaussian-p-posterior}
    p_{\theta} \left( \boldsymbol{x}_{t} \mid \boldsymbol{x}_{t+1} \right) \sim \mathcal{N} \left(\boldsymbol{\mu}_{\theta, t+1}(\boldsymbol{x}_{t+1}), \tilde{\sigma}_{t} \boldsymbol{I} \right),
\end{equation}
where
\begin{equation}
\label{eq:gaussian-p-posterior-param}
    \boldsymbol{\mu}_{\theta, t+1}(\boldsymbol{x}_{t+1}) = \frac{\sigma_{t}^{2}}{\sigma_{t+1}^{2}} \boldsymbol{x}_{t+1} + \frac{\sigma_{t+1}^{2} - \sigma_{t}^{2}}{\sigma_{t+1}^{2}} f \left(\boldsymbol{x}_{t+1}, t+1; \theta \right).
\end{equation}
Here, $f\left(\boldsymbol{x}_{t+1}, t+1; \theta \right)$ is a neural network with input of $(\boldsymbol{x}_{t+1}, t+1)$. Since $q \left( \boldsymbol{x}_{t} \mid \boldsymbol{x}_{0}, \boldsymbol{x}_{t+1} \right)$ and $p_{\theta} \left( \boldsymbol{x}_{t} \mid \boldsymbol{x}_{t+1} \right)$ have the same covariance matrix, we can derive that
\begin{equation}
\label{eq:guassian-kl}
\begin{split}
&D_{\mathrm{KL}} \left( q \left( \boldsymbol{x}_{t} \mid \boldsymbol{x}_{0}, \mathbf{x}_{t+1} \right)  \|  p_{\theta} \left( \boldsymbol{x}_{t} \mid \boldsymbol{x}_{t+1} \right)\right) \\
=& \left\| \boldsymbol{\mu}_{\theta, t+1}(\boldsymbol{x}_{t+1}) - \tilde{\boldsymbol{\mu}}_{t} \right\|_2^2.
\end{split}
\end{equation}
Neglecting the constant coefficient, minimizing \cref{eq:guassian-kl} is equivalent to minimize
\begin{equation}
\label{eq:simplified-guassian-kl}
\left\| f(\boldsymbol{x}_{t+1}, t+1; \theta) - \boldsymbol{x}_{0} \right\|_2^2
\end{equation}
Next, we turn to $\mathbb{E}_{q_{0,1,N}} \left[  - \log p_{\theta} \left(\boldsymbol{x}_{0} \mid \boldsymbol{x}_{1} \right) \right]$, the last term in \cref{eq:simplified-obj-assumption-1}. If we assume $p_{\theta} \left(\boldsymbol{x}_{0} \mid \boldsymbol{x}_{1} \right)$ follows some distribution, sampling from it may introduce extra undesired noise. Hence, practically we replace the sampling by $\mathbb{E} \left[ \boldsymbol{x}_{0} \mid \boldsymbol{x}_{1} \right]$. As a result, we train $p_{\theta}$ by $\mathbb{E}_{q_{0,1,N}} \left[ \left\| f\left(\boldsymbol{x}_{1}, 1; \theta \right) - \boldsymbol{x}_{0} \right\|_2^2 \right]$ instead of $\mathbb{E}_{q_{0,1,N}} \left[  - \log p_{\theta} \left(\boldsymbol{x}_{0} \mid \boldsymbol{x}_{1} \right) \right]$.

Combining with the above analysis, we give the final objective loss function as follows:
\begin{equation}
\label{eq:gaussian-final-loss}
L = \mathbb{E}_{q} \sum_{t = 0}^{t=N-1} \left\| f(\boldsymbol{x}_{t+1}, t+1; \theta) - \boldsymbol{x}_{0} \right\|_2^2.
\end{equation}

At last, we show the full training and sampling algorithms in \cref{alg:gaussian-training} and \cref{alg:gaussian-sampling}.

\begin{algorithm}[t]
\renewcommand{\algorithmicrequire}{\textbf{Input:}}
\renewcommand{\algorithmicensure}{\textbf{Output:}}
\caption{The training process for Gaussian noise}
\label{alg:gaussian-training}
\begin{algorithmic}[1]
    \REQUIRE $\left\{ \boldsymbol{x}_{0} \right\}$, $\left\{ \sigma_{t} \right\}$, $f \left(\cdot, \cdot, \theta\right)$.
    \ENSURE trained $f \left(\cdot, \cdot, \theta\right)$.
    \WHILE{$\theta$ is not converged}
        \STATE Random select $\boldsymbol{x}_0$ and sample $t$ from $\left\{1, ..., N \right\}$ uniformly.
        \STATE Sample $\boldsymbol{x}_{t}$ from $\mathcal{N}(\boldsymbol{x}_0, \sigma^2_{t} \boldsymbol{I})$.
        \STATE Compute $\mathrm{grad}$ by $\nabla_{\theta} \left\| f(\boldsymbol{x}_{t}, t; \theta) - \boldsymbol{x}_{0} \right\|_2^2$.
        \STATE Update $\theta$ by $\mathrm{grad}$.
    \ENDWHILE
\end{algorithmic}  
\end{algorithm}

\begin{algorithm}[t]
\renewcommand{\algorithmicrequire}{\textbf{Input:}}
\renewcommand{\algorithmicensure}{\textbf{Output:}}
\caption{The sampling process for Gaussian noise}
\label{alg:gaussian-sampling}
\begin{algorithmic}[1]
    \REQUIRE noisy image $\boldsymbol{x}_{N}$ and trained $f \left(\cdot, \cdot, \theta\right)$.
    \ENSURE ${\boldsymbol{x}}_{0}$.
    \FOR{$t = N-1, ..., 1$}
        \STATE Sample $\boldsymbol{x}_t$ by 
        \begin{equation}
        \mathcal{N} \left( \frac{\sigma_{t}^{2}}{\sigma_{t+1}^{2}} \boldsymbol{x}_{t+1} + \frac{\sigma_{t+1}^{2} - \sigma_{t}^{2}}{\sigma_{t+1}^{2}} f \left(\boldsymbol{x}_{t+1}, t+1; \theta \right), \tilde{\sigma}_{t} \boldsymbol{I} \right).
        \end{equation}
    \ENDFOR
    \STATE $\boldsymbol{x}_{0} = f \left(\boldsymbol{x}_{1}, 1, \theta\right)$
\end{algorithmic}  
\end{algorithm}

\subsubsection{Gamma Noise}
\label{sec:gamma}

The Gamma noise model in denoising task is defined by the following form:
\begin{equation}
    \boldsymbol{x}_{N} = \boldsymbol{\eta} \odot \boldsymbol{x}_{0}, \eta_i \sim \frac{1}{\alpha} \mathcal{G} \left(\alpha, 1 \right),
\end{equation}
where $ \alpha > 1$ and $\mathcal{G} \left(\alpha, 1 \right)$ is a Gamma distribution with parameters of $\alpha$ and $1$. $\odot$ represents component-wise multiplication. For the sake of convenience, we neglect it in notation if not ambiguous.

We select $\alpha_{0} = \infty > \alpha_{1} > \cdots > \alpha_{N} = \alpha$. Therefore, $\left\{ \alpha_{t} \right\}$ is a monotonically decreasing sequence. Let
\begin{equation}
\label{eq:gamma-assumption-eq1}
\boldsymbol{x}_{1} = \boldsymbol{\eta}_{1} \boldsymbol{x}_{0}, \quad \eta_{1, i} \sim \frac{1}{\alpha_{1}}  \mathcal{G} \left( \alpha_{1}, 1 \right),
\end{equation}
and 
\begin{equation}
\label{eq:gamma-assumption-eq2}
\boldsymbol{x}_{t+1} = \frac{\alpha_{t}}{\alpha_{t+1}} \boldsymbol{\zeta}_{t+1} \boldsymbol{x}_{t}, \zeta_{t+1, i} \sim \mathcal{B} \left( \alpha_{t+1}, \alpha_{t} - \alpha_{t+1} \right)
\end{equation}
where $t = 1, ..., N-1$ and $\mathcal{B} \left( \alpha_{t+1}, \alpha_{t} - \alpha_{t+1} \right)$ is a Beta distribution with parameters of $\alpha_{t+1}$ and $\alpha_{t} - \alpha_{t+1}$. Then we have:
\begin{equation}
\label{eq:gamma-xt}
\begin{split}
\boldsymbol{x}_{t} = \boldsymbol{\eta}_{t} \boldsymbol{x}_{0}, \quad \eta_{t, i} \sim \frac{1}{\alpha_{t}}  \mathcal{G} \left( \alpha_{t}, 1 \right),
\end{split}
\end{equation}
where $t = 0, 1, ..., N-1$. The proof of \cref{eq:gamma-xt} is in Appendix. \Cref{eq:gamma-assumption-eq2} define the relation between $\mathbf{x}_{t}$, $t = 1, ..., N$. By now, we define a full diffusion process for Gamma Noise. Obviously, \cref{eq:assumption-1} holds according to \cref{eq:gamma-assumption-eq2}. Thus, $\mathbf{x}_{t}$, $t = 1, ..., N$ are also a Markov chain.

Based on the analysis in \cref{sec:gaussian}, we know that all the equations from \cref{eq:property-1-assumption-1} to \cref{eq:simplified-obj-assumption-1} still hold.

Now, we consider the main part of the loss function $\mathbb{E}_{q_{0,t+1, N}} \left[ D_{\mathrm{KL}} \left( q \left( \boldsymbol{x}_{t} \mid \boldsymbol{x}_{0}, \boldsymbol{x}_{t+1} \right) \| p_{\theta} \left( \boldsymbol{x}_{t} \mid \boldsymbol{x}_{t+1} \right)  \right) \right]$. We begin with the analytical form of $q \left( \boldsymbol{x}_{t} \mid \boldsymbol{x}_{0}, \boldsymbol{x}_{t+1} \right)$. We have the conclusion that:
\begin{equation}
\label{eq:gamma-q-posterior-eq1}
\begin{split}
    \left(\frac{\alpha_{t} \boldsymbol{x}_{t} - \alpha_{t+1} \boldsymbol{x}_{t+1}}{\boldsymbol{x}_{0}}\right)_{i} \sim \mathcal{G} \left(\alpha_{t} - \alpha_{t+1}, 1 \right), t = 1, ..., N-1.
\end{split}
\end{equation}
Here, the division is component-wise operation. Thus, $q \left( \boldsymbol{x}_{t} \mid \boldsymbol{x}_{0}, \boldsymbol{x}_{t+1} \right)$ can be represented by
\begin{equation}
\label{eq:gamma-q-posterior-eq2}
\begin{split}
    \boldsymbol{x}_{t} = \frac{\boldsymbol{x}_{0} \boldsymbol{\tau}_{t} + \alpha_{t+1} \boldsymbol{x}_{t+1}}{\alpha_{t}}, \tau_{t, i} \sim \mathcal{G} \left(\alpha_{t} - \alpha_{t+1}, 1 \right).
\end{split}
\end{equation}
Therefore, we assume that $p_{\theta} \left( \boldsymbol{x}_{t} \mid \boldsymbol{x}_{t+1} \right)$, $t = 1, ..., N-1$ has the following form:
\begin{equation}
\label{eq:gamma-p-posterior}
    \boldsymbol{x}_{t} = \frac{f\left(\boldsymbol{x}_{t+1}, t+1; \theta \right) \boldsymbol{\tau}_{t} + \alpha_{t+1} \boldsymbol{x}_{t+1}}{\alpha_{t}}, {\tau}_{t, i} \sim \mathcal{G} (\alpha_{t} - \alpha_{t+1}, 1)
\end{equation}
where $f\left(\boldsymbol{x}_{t+1}, t+1; \theta \right)$ is a neural network with input of $(\boldsymbol{x}_{t+1}, t+1)$. Then we can derive that
\begin{equation}
\label{eq:gamma-kl}
\begin{split}
&D_{\mathrm{KL}} \left( q \left( \boldsymbol{x}_{t} \mid \boldsymbol{x}_{0}, \mathbf{x}_{t+1} \right)  \|  p_{\theta} \left( \boldsymbol{x}_{t} \mid \boldsymbol{x}_{t+1} \right)\right) \\
=& \sum_{i} \left(\alpha_{t} - \alpha_{t+1} \right)\left( \log \frac{f_{\theta, i}}{\boldsymbol{x}_{0}} + \frac{\boldsymbol{x}_{0}}{f_{\theta, i}} - 1  \right).
\end{split}
\end{equation}
Here, $f_{\theta}$ is the abbreviation of $f\left(\boldsymbol{x}_{t+1}, t+1; \theta \right)$. Suppose $f_{\theta^*}$ is the optimal function minimizing \cref{eq:gamma-kl}, we can prove that it is also the optimal function for the following optimization problem:
\begin{equation}
\min_{f_{\theta}} \mathbb{E}_{q} \left\|f(\boldsymbol{x}_{t+1}, t+1; \theta) - \boldsymbol{x}_{0} \right\|_2^2.
\end{equation}
The proof is in Appendix. Hence, training $f_{\theta}$ by minimize KL divergence is equivalent to train $f_{\theta}$ by L2 norm loss function with $\boldsymbol{x}_{0}$ as labels.

As for $\mathbb{E}_{q_{0,1,N}} \left[  - \log p_{\theta} \left(\boldsymbol{x}_{0} \mid \boldsymbol{x}_{1} \right) \right]$, the last term in \cref{eq:simplified-obj-assumption-1}, we adapt the same strategy described in \cref{sec:gaussian}. As a result, the final objective loss function is as follows:
\begin{equation}
\label{eq:gamma-final-loss}
L = \mathbb{E}_{q} \sum_{t = 0}^{t=N-1} \left\| f(\boldsymbol{x}_{t+1}, t+1; \theta) - \boldsymbol{x}_{0} \right\|_2^2.
\end{equation}

At last, we show the full training and sampling algorithms in \cref{alg:gamma-training} and \cref{alg:gamma-sampling}.

\begin{algorithm}[t]
\renewcommand{\algorithmicrequire}{\textbf{Input:}}
\renewcommand{\algorithmicensure}{\textbf{Output:}}
\caption{The training process for Gamma noise}
\label{alg:gamma-training}
\begin{algorithmic}[1]
    \REQUIRE $\left\{ \boldsymbol{x}_{0} \right\}$, $\left\{ \alpha_{t} \right\}$, $f \left(\cdot, \cdot, \theta\right)$.
    \ENSURE trained $f \left(\cdot, \cdot, \theta\right)$.
    \WHILE{$\theta$ is not converged}
        \STATE Random select $\boldsymbol{x}_0$ and sample $t$ from $\left\{1, ..., N \right\}$ uniformly.
        \STATE Sample $\boldsymbol{\eta}_{t}$ from $\mathcal{G} \left( \alpha_{t}, 1 \right)$
        \STATE $\boldsymbol{x}_{t} = \boldsymbol{\eta}_{t}\boldsymbol{x}_{t}$ .
        \STATE Compute $\mathrm{grad}$ by $\nabla_{\theta} \left\| f(\boldsymbol{x}_{t}, t; \theta) - \boldsymbol{x}_{0} \right\|_2^2$.
        \STATE Update $\theta$ by $\mathrm{grad}$.
    \ENDWHILE
\end{algorithmic}  
\end{algorithm}

\begin{algorithm}[t]
\renewcommand{\algorithmicrequire}{\textbf{Input:}}
\renewcommand{\algorithmicensure}{\textbf{Output:}}
\caption{The sampling process for Gamma noise}
\label{alg:gamma-sampling}
\begin{algorithmic}[1]
    \REQUIRE noisy image $\boldsymbol{x}_{N}$, $\left\{ \alpha_{t} \right\}$, and trained $f \left(\cdot, \cdot, \theta\right)$.
    \ENSURE ${\boldsymbol{x}}_{0}$.
    \FOR{$t = N-1, ..., 1$}
        \STATE Sample $\boldsymbol{\tau}_t$ from $\mathcal{G} (\alpha_{t} - \alpha_{t+1}, 1)$.
        \STATE $\boldsymbol{x}_{t} = \frac{1}{\alpha_{t}}\left(f(\boldsymbol{x}_{t+1}, t+1; \theta) \boldsymbol{\tau}_{t} + \alpha_{t+1} \boldsymbol{x}_{t+1}\right)$
    \ENDFOR
    \STATE $\boldsymbol{x}_{0} = f \left(\boldsymbol{x}_{1}, 1, \theta\right)$
\end{algorithmic}  
\end{algorithm}

\subsubsection{Poisson Noise}
\label{sec:poisson}

The Poisson noise model in denoising task is defined by the following form:
\begin{equation}
    \boldsymbol{x}_{N} = \frac{ \mathcal{P}\left( \lambda \boldsymbol{x}_{0} \right)}{\lambda},
\end{equation}
where $ \lambda > 0$ and $\mathcal{P}\left( \lambda \boldsymbol{x}_{0} \right)$ is a Poisson distribution with parameters of $\lambda \boldsymbol{x}_{0}$. 

We select $\infty = \lambda_{0} > \lambda_{1} > \cdots > \lambda_{N} = \lambda$. Therefore, $\left\{ \lambda_{t} \right\}$ is a monotonically decreasing sequence. Let
\begin{equation}
\label{eq:poisson-assumption-eq1}
\boldsymbol{x}_{N} \mid \boldsymbol{x}_{0} \sim \frac{\mathcal{P} \left( \lambda_{N} \boldsymbol{x}_{0} \right)}{\lambda_{N}},
\end{equation}
and 
\begin{equation}
\label{eq:poisson-assumption-eq2}
\boldsymbol{x}_{t} \mid \boldsymbol{x}_{t+1}, \boldsymbol{x}_{0} \sim \frac{\lambda_{t+1} \boldsymbol{x}_{t+1} + \mathcal{P} \left( \left(\lambda_{t} - \lambda_{t+1}\right) \boldsymbol{x}_{0} \right)}{\lambda_{t}},
\end{equation}
where $t = 1, ..., N-1$. Then we have:
\begin{equation}
\label{eq:poisson-xt}
\begin{split}
\boldsymbol{x}_{t} \mid \boldsymbol{x}_{0} \sim \frac{\mathcal{P}  \left( \lambda_{t} \boldsymbol{x}_{0} \right)}{\lambda_{t}},
\end{split}
\end{equation}
where $t = 1, ..., N-1$. The proof of \cref{eq:poisson-xt} is in Appendix. \Cref{eq:poisson-assumption-eq2} define the relation between $\mathbf{x}_{t}$, $t = 1, ..., N$. by now, we define a full diffusion process for Poison Noise. According to \cref{eq:poisson-assumption-eq2}, we have
\begin{equation}
\label{eq:assumption-2}
    q \left( \boldsymbol{x}_{t} \mid \boldsymbol{x}_{0}, \boldsymbol{x}_{t+1:N} \right) = q \left( \boldsymbol{x}_{t} \mid \boldsymbol{x}_{0}, \boldsymbol{x}_{t+1} \right).
\end{equation}
From \cref{eq:assumption-2}, we can derive that
\begin{equation}
\label{eq:property-1-assumption-2}
\begin{split}
    q \left( \boldsymbol{x}_{1:N-1} \mid \boldsymbol{x}_{0}, \boldsymbol{x}_{N} \right)
    = & \prod_{t=1}^{N-1} q \left( \boldsymbol{x}_{t} \mid \boldsymbol{x}_{0}, \boldsymbol{x}_{t+1:N} \right) \\
    = & \prod_{t=1}^{N-1} q \left( \boldsymbol{x}_{t} \mid \boldsymbol{x}_{0}, \boldsymbol{x}_{t+1}\right).
\end{split}
\end{equation}
Applying \cref{eq:property-1-assumption-2} to \cref{eq:general-loss}, we can derive the same result as \cref{eq:obj-assumption-1}:
\begin{equation}
\label{eq:obj-assumption-1-v2}
\begin{split}
    L & \leq \sum_{t=1}^{N-1} \mathbb{E}_{q_{0,t+1, N}} \left[ D_{\mathrm{KL}} \left( q \left( \boldsymbol{x}_{t} \mid \boldsymbol{x}_{0}, \boldsymbol{x}_{t+1} \right) \| p_{\theta} \left( \boldsymbol{x}_{t} \mid \boldsymbol{x}_{t+1}, \mathbf{x}_{N} \right)  \right) \right] \\
    & \quad \quad +  \mathbb{E}_{q_{0,1,N}} \left[  - \log p_{\theta} \left(\boldsymbol{x}_{0} \mid \boldsymbol{x}_{1}, \boldsymbol{x}_{N} \right) \right].
\end{split}
\end{equation}
However, $\boldsymbol{x}_{N}$ in $p_{\theta} \left( \boldsymbol{x}_{t} \mid \boldsymbol{x}_{t+1}, \boldsymbol{x}_{N} \right)$ cannot be removed. Thus, \cref{eq:simplified-obj-assumption-1} does not hold for Poisson noise model.

Now, we consider the main part of the loss function $\mathbb{E}_{q_{0,t+1, N}} \left[ D_{\mathrm{KL}} \left( q \left( \boldsymbol{x}_{t} \mid \boldsymbol{x}_{0}, \boldsymbol{x}_{t+1} \right) \| p_{\theta} \left( \boldsymbol{x}_{t} \mid \boldsymbol{x}_{t+1} \right)  \right) \right]$. We have known the analytical form of $q \left( \boldsymbol{x}_{t} \mid \boldsymbol{x}_{0}, \boldsymbol{x}_{t+1} \right)$ from \cref{eq:poisson-assumption-eq2}. Therefore, we assume that $p_{\theta} \left( \boldsymbol{x}_{t} \mid \boldsymbol{x}_{t+1}, \boldsymbol{x}_{0} \right)$, $t = 1, ..., N-1$ has the following form:
\begin{equation}
\label{eq:poisson-p-posterior}
\begin{split}
    & p_{\theta} \left( \boldsymbol{x}_{t} \mid \boldsymbol{x}_{t+1}, \boldsymbol{x}_{N} \right) \\
    \sim & \frac{\lambda_{t+1} \boldsymbol{x}_{t+1} + \mathcal{P} \left( \left(\lambda_{t} - \lambda_{t+1}\right) f(\boldsymbol{x}_{t+1}, \boldsymbol{x}_{N}, t+1; \theta) \right)}{\lambda_{t}},
\end{split}
\end{equation}
where $f(\boldsymbol{x}_{t+1}, \boldsymbol{x}_{N}, t+1; \theta)$ is a neural network with input of $(\boldsymbol{x}_{t+1},\boldsymbol{x}_{N}, t+1)$. Denote $f(\boldsymbol{x}_{t+1}, \boldsymbol{x}_{N}, t+1; \theta)$ as $f_{\theta}$ for simplicity, then we can derive that
\begin{equation}
\label{eq:poisson-kl}
\begin{split}
&D_{\mathrm{KL}} \left( q \left( \boldsymbol{x}_{t} \mid \boldsymbol{x}_{0}, \mathbf{x}_{t+1} \right)  \|  p_{\theta} \left( \boldsymbol{x}_{t} \mid \boldsymbol{x}_{t+1}, \boldsymbol{x}_{N} \right)\right) \\
=& \left( \log \boldsymbol{x}_{0} - \log f_{\theta} \right) \left( \lambda_{t} - \lambda_{t+1} \right) \mathbf{x}_{0} \\
& \quad \quad - \left( \lambda_{t} - \lambda_{t+1} \right) \left(\boldsymbol{x}_{0} - f_{\theta} \right).
\end{split}
\end{equation}
Similar to the analysis in \cref{sec:gamma}, we attempt to transform the original optimization problem to another equivalent one. Suppose $f_{\theta^*}$ is the optimal function minimizing \cref{eq:poisson-kl}, we can prove that it is also the optimal function for the following optimization problem:
\begin{equation}
\label{eq:poisson-mini}
\min_{f_{\theta}} \mathbb{E}_{q} \left\|f(\boldsymbol{x}_{t+1}, \boldsymbol{x}_{N}, t+1; \theta) - \boldsymbol{x}_{0} \right\|_2^2.
\end{equation}
The proof is in Appendix. Hence, training $f_{\theta}$ by minimize KL divergence is equivalent to train $f_{\theta}$ by L2 norm loss function with $\boldsymbol{x}_{0}$ as labels.

As for $\mathbb{E}_{q_{0,1,N}} \left[  - \log p_{\theta} \left(\boldsymbol{x}_{0} \mid \boldsymbol{x}_{1}, \boldsymbol{x}_{N} \right) \right]$, the last term in \cref{eq:obj-assumption-1-v2}, we still adapt the same strategy described in \cref{sec:gaussian}. As a result, the final objective loss function is as follows:
\begin{equation}
\label{eq:poisson-final-loss}
L = \mathbb{E}_{q} \sum_{t = 0}^{t=N-1} \left\| f(\boldsymbol{x}_{t+1}, \boldsymbol{x}_{N}, t+1; \theta) - \boldsymbol{x}_{0} \right\|_2^2.
\end{equation}

At last, we show the full training and sampling algorithms in \cref{alg:poisson-training} and \cref{alg:poisson-sampling}.

\begin{algorithm}[t]
\renewcommand{\algorithmicrequire}{\textbf{Input:}}
\renewcommand{\algorithmicensure}{\textbf{Output:}}
\caption{The training process for Poisson noise}
\label{alg:poisson-training}
\begin{algorithmic}[1]
    \REQUIRE $\left\{ \boldsymbol{x}_{0} \right\}$, $\left\{ \lambda_{t} \right\}$, $f \left(\cdot, \cdot, \theta\right)$.
    \ENSURE trained $f \left(\cdot, \cdot, \theta\right)$.
    \WHILE{$\theta$ is not converged}
        \STATE Random select $\boldsymbol{x}_0$ and sample $t$ from $\left\{1, ..., N-1 \right\}$ uniformly.
        \STATE Sample $\boldsymbol{x}_{N}$ from $\frac{\mathcal{P} \left( \lambda_{N} \mathbf{x}_{0} \right)}{\lambda_{N}}$.
        \STATE Sample $\boldsymbol{x}_{t}$ from $\frac{1}{\lambda_{t}} \left(\lambda_{t+1} \mathbf{x}_{t+1} + \mathcal{P} \left( \left(\lambda_{t} - \lambda_{t+1}\right) \mathbf{x}_{0} \right)\right)$
        \STATE Compute $\mathrm{grad}$ by $\nabla_{\theta} \left\| f(\boldsymbol{x}_{t}, \boldsymbol{x}_{N}, t; \theta) - \boldsymbol{x}_{0} \right\|_2^2$.
        \STATE Update $\theta$ by $\mathrm{grad}$.
    \ENDWHILE
\end{algorithmic}  
\end{algorithm}

\begin{algorithm}[t]
\renewcommand{\algorithmicrequire}{\textbf{Input:}}
\renewcommand{\algorithmicensure}{\textbf{Output:}}
\caption{The sampling process for Poisson noise}
\label{alg:poisson-sampling}
\begin{algorithmic}[1]
    \REQUIRE noisy image $\boldsymbol{x}_{N}$, $\left\{ \lambda_{t} \right\}$ and trained $f \left(\cdot, \cdot, \theta\right)$.
    \ENSURE ${\boldsymbol{x}}_{0}$.
    \FOR{$t = N-1, ..., 1$}
        \STATE Sample $\boldsymbol{\tau}_t$ from $\mathcal{P} \left( \left(\lambda_{t} - \lambda_{t+1}\right) f(\mathbf{x}_{t+1}, \mathbf{x}_{N}, t; \theta) \right)$ 
        \STATE $\boldsymbol{x}_{t} = \frac{1}{\lambda_{t}} \left(\lambda_{t+1} \boldsymbol{x}_{t+1} + \boldsymbol{\tau}_t\right)$
    \ENDFOR
    \STATE $\boldsymbol{x}_{0} = f \left(\boldsymbol{x}_{1}, \boldsymbol{x}_{N}, 1, \theta\right)$
\end{algorithmic}  
\end{algorithm}

\subsection{Discussion}
\label{sec:discussion}

In this section, we have discussed three types of noise models in the denoising task, Gaussian, Gamma and Poisson noise. The diffusion process of Gaussian and Gamma noise can be defined as a Markov chain, representing the evolution from clean images to noisy images. Gaussian distribution itself is additive. Thus, its diffusion process is only related to Gaussian distribution. While Beta distribution is introduced to define the diffusion process for Gamma noise. As for Poisson noise, the difference is clear. From definition, its diffusion process can also be regarded as another form of Markov chain, which is conditioned on $\boldsymbol{x}_{0}$ and represents an evolution from noisier images to less noisy images. Hence, the definition of diffusion process is related to the statistical property of the noise model. 

About the derivation of objective loss functions of model training, the basic idea is to minimize KL divergence. In the case of Gamma noise and Poisson noise, we transform the original complex optimization problem to L2 norm loss minimization through optimization equivalence. As a result, the model training for three noise models are highly consistent. Though the input of models are slightly different, they can all be written as
\begin{equation}
\label{eq:loss-consistent}
\min \mathbb{E}_{q, t} \left\|f_{\theta, t} - \boldsymbol{x}_{0} \right\|_2^2.
\end{equation}
In other words, given $t$ the model training amounts to estimate the posterior mean, $\mathbb{E} \left[\boldsymbol{x}_{0} \mid \boldsymbol{x}_{t+1}, \boldsymbol{x}_{N} \right]$. Therefore, $p_{\theta} \left(\boldsymbol{x}_{t} \mid \boldsymbol{x}_{t+1}, \boldsymbol{x}_{N} \right)$ is constructed through replacing the $\boldsymbol{x}_{0}$ in $q \left( \boldsymbol{x}_{t} \mid \boldsymbol{x}_{0}, \boldsymbol{x}_{t+1}\right)$ by the posterior mean.

\section{Experiment}
\label{sec:experiment}


\begin{figure*}
  \centering
  \includegraphics[width=0.85\linewidth]{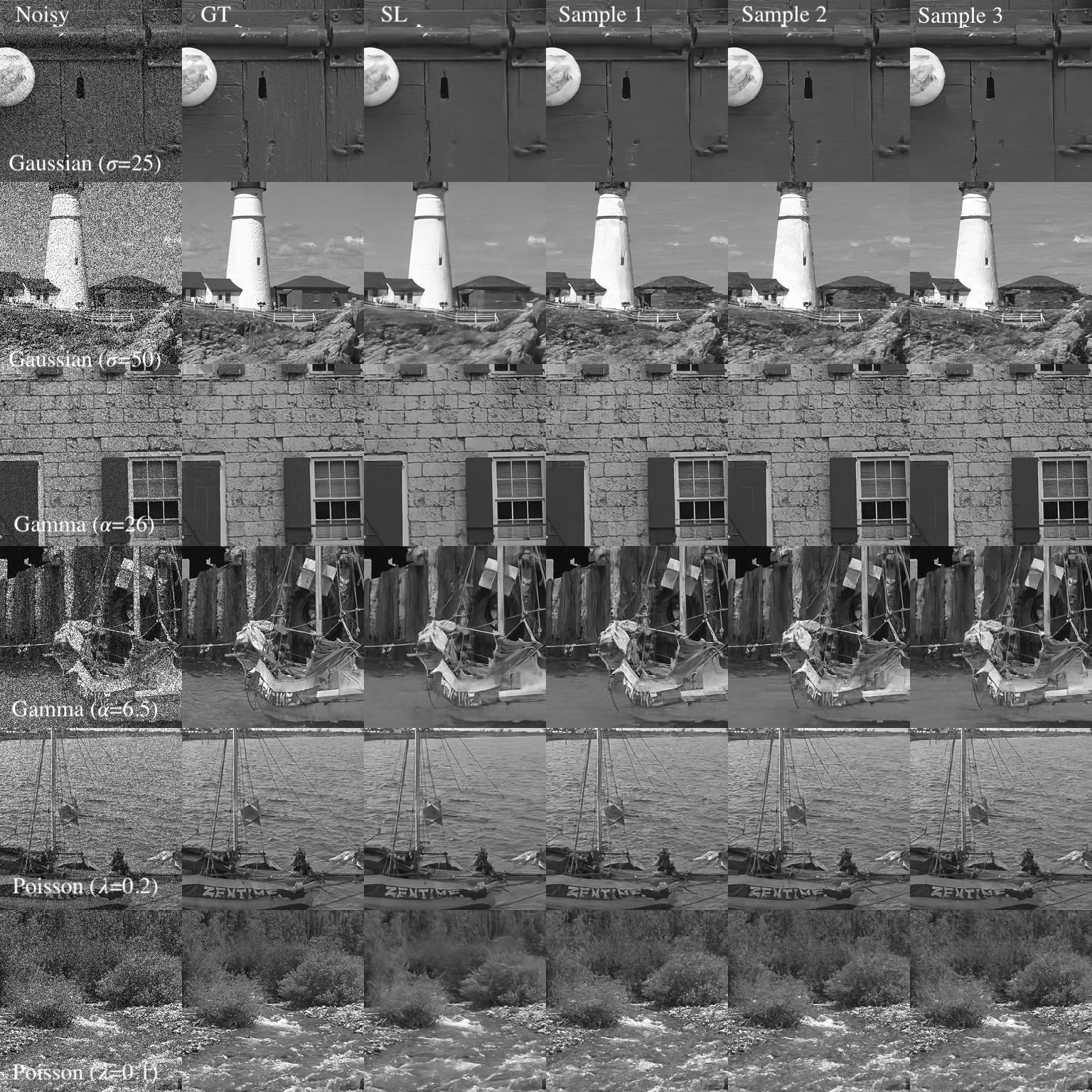}
  \caption{Qualitative comparison using Kodak dataset. From top to bottom is: Gaussian noise with $\sigma=25$ and $\sigma=50$, Gamma noise with $\alpha=26$ and $\alpha=6.5$, and Poisson noise with $\lambda=0.2$ and $\lambda=0.1$. Noisy: noisy image, GT: ground-truth, SL: supervised learning. The last three columns are generated different samples.}
  \label{fig:results}
\end{figure*}

\begin{table*}
  \centering
  \begin{tabular}{@{}l|ccc|ccc@{}}
    \toprule
    Noise Model & \multicolumn{3}{c}{Kodak}\vline & \multicolumn{3}{c}{CSet9} \\
    \hline
     & SL & Samples & Mean of Samples & SL & Samples & Mean of Samples \\
     \hline
    Gaussian, $\sigma = 25$
       & 30.53 / 0.816 & 28.38 / 0.755 &  30.45 / 0.820 & 29.41 / 0.843 & 27.35 / 0.792 &  29.41 / 0.849 \\
    Gaussian, $\sigma = 50$
       & 27.69 / 0.742 & 25.30 / 0.645 &  27.63 / 0.742 & 26.11 / 0.760 & 23.69 / 0.664 &  26.26 / 0.766 \\

    Gamma, $\alpha = 26$
       & 31.18 / 0.849 & 28.59 / 0.789 &  30.84 / 0.852 & 29.68 / 0.854 & 27.55 / 0.807 &  29.64 / 0.859 \\
    Gamma, $\alpha = 6.5$
       & 28.31 / 0.776 & 26.40 / 0.710 &  28.23 / 0.775 & 26.52 / 0.776 & 24.51 / 0.712 &  26.62 / 0.781 \\
       
    Poisson, $\lambda = 0.2$
       & 30.95 / 0.834 & 28.71 / 0.770 &  30.76 / 0.834 & 29.68 / 0.851 & 27.53 / 0.782 &  29.69 / 0.851 \\
    Poisson, $\lambda = 0.1$
       & 29.31 / 0.794 & 26.96 / 0.696 &  29.28 / 0.792 & 27.88 / 0.810 & 25.53 / 0.706 &  27.97 / 0.810 \\
    \bottomrule
  \end{tabular}
  \caption{Quantitative comparison in terms of PNSR (dB) / SSIM for different noise model on Kodak and CSet9 datasets. We compare supervised learning (SL) to our method. "Samples" means the generated results while "Mean of Samples" means the average of $100$ samples for each noisy images. }
  \label{tab:result-metric}
\end{table*}

We conduct extensive experiments to evaluate our approach, including Gaussian noise, Gamma noise and Poisson noise with different noise levels. 

\paragraph{Dataset and Implementation Details} We evaluate the proposed method for gray images in the two benchmark datasets: KOdak dataset and CSet9. DIV2K \cite{timofte2018ntire} and CBSD500 dataset \cite{chaudhary2019comparative} are used as training datasets. Original images are color RGB natural images and we transform them to gray images when training and testing. We use traditional supervised learning (\cref{eq:sl-training}) with as L2 norm loss as the baseline for comparison. The same modified U-Net \cite{dhariwal2021diffusion} with about 70 million parameters is used for all methods. When training, we randomly clip the training images to patches with the resolution of $128 \times 128$. AdamW optimizer \cite{loshchilov2017decoupled} is used to train the network. We train each model with the batch size of 32. To reduce memory, we utilize the tricks of cumulative gradient and mixed precision training. The learning rate is set as $1 \times 10^{-4}$. All the models are implemented in PyTorch \cite{paszke2017automatic} with NVidia V100. The pixel value range of all clean images are $[0, 255]$ and the parameters of noise models are built on it. When training models, images will be scaled to $[-1, 1]$. After generating samples, they will be scaled back to $[0, 255]$. For each type of noise model, we choose two noise level and different number of diffusion steps, $N$. 
We list them here:
\begin{itemize}
    \item Gaussian: $
    \sigma = 25$ ($N = 20$), $
    \sigma = 50$ ($N = 40$);
    \item Gamma: $
    \alpha = 26$ ($N = 20$), $
    \alpha = 6.5$ ($N = 40$);
    \item Poisson: $
    \lambda = 0.2$ ($N = 20$), $
    \lambda = 0.1$ ($N = 40$).
\end{itemize}
Another setting which is not discussed in \cref{sec:method} is how to construct the sequence of noise model parameters. In our experiments, we adapt a simple but effective way in which the sequence is constructed such that the standard deviation of $\mathbf{x}_{t} \mid \mathbf{x}_{0}$ is linearly increased from $t = 0$ to $t = N$. The more details of implementation are described in Appendix.

\paragraph{Results}

\Cref{fig:results} shows the generated samples for different noise models. Compared to supervised learning, our generated results are visually pleasing, containing better image details. The good visual quality of generated samples verifies that our method is feasible for generative denoising tasks. We also compute the PSNR and SSIM to evaluate our method as shown in \cref{tab:result-metric}. We discover that there is a gap between supervised learning and generated samples in terms of metrics. This is understandable. Because noisy images have lost much original image information, it is hard to generate completely identical details to those lost ones. For each noisy images, we generate 100 samples and compute the mean of them as the estimation of $p \left(\boldsymbol{x} \mid \boldsymbol{y} \right)$, which are also illustrated in \cref{tab:result-metric}. Apparently, the metrics of sample mean is close to supervised learning, which further verifies that the posterior distribution estimated by our method is effective. We also investigate the effect of the number of steps, $N$. Due to the limit of paper length, we discuss it in Appendix.


\section{Conclusion}
\label{sec:conclusion}
 
In this paper, we apply the framework of diffusion models to generative image denoising tasks and propose a new diffusion model based on the image noise model. Different to the original diffusion model, we redefine the diffusion process according to the specific noise models and derive the model training and sampling algorithms. Interestingly, we find that model training for Gaussian, Gamma and Poisson noise is unified to a highly consistent strategy. Our experiments verify that our method is feasible for generative image denoising. In the future, we hope to extend our method to other noise model and evaluate its performance on other dataset.



\appendix

\section{Proofs}
\label{sec:proof}

\subsection{The Proof of \cref{eq:general-loss} in Sec. 3.1.3}

\begin{equation}
\begin{split}
L \leq \mathbb{E}_{q_{0:N}}\left[- \log \frac{\prod_{t=0}^{t=N-1} p_{\theta} \left( \boldsymbol{x}_{t} \mid \boldsymbol{x}_{t+1}, \boldsymbol{x}_{N} \right)  }{q \left( \boldsymbol{x}_{1:N-1} \mid \boldsymbol{x}_{0}, \boldsymbol{x}_{N} \right)}\right].
\end{split}
\tag{10}
\end{equation}
The full derivation is as follows.
\begin{proof}
\begin{align*}
L &= \mathbb{E}_{\boldsymbol{x}_{0}, \boldsymbol{x}_{N}} \left[ - \log p_{\theta}\left( \boldsymbol{x}_{0} \mid \boldsymbol{x}_{N} \right) \right] \\
&= \mathbb{E}_{\boldsymbol{x}_{0}, \boldsymbol{x}_{N}} \left[ - \log \int_{\boldsymbol{X}_{1}, ..., \boldsymbol{X}_{N-1}} p_{\theta}\left( \boldsymbol{x}_{0}, \boldsymbol{x}_{1}, ..., \boldsymbol{x}_{N-1} \mid \boldsymbol{x}_{N} \right) \right. \\
& \quad\quad \quad \left. \mathrm{d} \boldsymbol{x}_{1} \cdots \mathrm{d} \boldsymbol{x}_{N-1} \right] \\
&= \mathbb{E}_{\boldsymbol{x}_{0}, \boldsymbol{x}_{N}} \left[ - \log \int_{\boldsymbol{X}_{1}, ..., \boldsymbol{X}_{N-1}} \frac{p_{\theta}\left( \boldsymbol{x}_{0}, \boldsymbol{x}_{1}, ..., \boldsymbol{x}_{N-1} \mid \boldsymbol{x}_{N} \right) }{q \left( \boldsymbol{x}_{1}, ..., \boldsymbol{x}_{N-1} \mid \boldsymbol{x}_{0}, \boldsymbol{x}_{N} \right)} \right. \\
& \quad\quad \quad \left. q \left( \boldsymbol{x}_{1}, ..., \boldsymbol{x}_{N-1} \mid \boldsymbol{x}_{0}, \boldsymbol{x}_{N} \right) \mathrm{d} \boldsymbol{x}_{1} \cdots \mathrm{d} \boldsymbol{x}_{N-1}\right] \\
&\leq \mathbb{E}_{\boldsymbol{x}_{0}, \boldsymbol{x}_{N}} \left[ - \int_{\boldsymbol{X}_{1}, ..., \boldsymbol{X}_{N-1}} q \left( \boldsymbol{x}_{1}, ..., \boldsymbol{x}_{N-1} \mid \boldsymbol{x}_{0}, \boldsymbol{x}_{N} \right) \right. \\
& \quad\quad \quad \left. \log \frac{p_{\theta}\left( \boldsymbol{x}_{0}, \boldsymbol{x}_{1}, ..., \boldsymbol{x}_{N-1} \mid \boldsymbol{x}_{N} \right) }{q \left( \boldsymbol{x}_{1}, ..., \boldsymbol{x}_{N-1} \mid \boldsymbol{x}_{0}, \boldsymbol{x}_{N} \right)}   \mathrm{d} \boldsymbol{x}_{1} \cdots \mathrm{d} \boldsymbol{x}_{N-1}\right] \\
& = \mathbb{E}_{\boldsymbol{x}_{0}, \boldsymbol{x}_{N}} \left[ \mathbb{E}_{\boldsymbol{x}_{1}, ..., \boldsymbol{x}_{N-1} \mid \boldsymbol{x}_{0}, \boldsymbol{x}_{N}} \left[- \log \frac{p_{\theta}\left( \boldsymbol{x}_{0}, \boldsymbol{x}_{1}, ..., \boldsymbol{x}_{N-1} \mid \boldsymbol{x}_{N} \right) }{q \left( \boldsymbol{x}_{1}, ..., \boldsymbol{x}_{N-1} \mid \boldsymbol{x}_{0}, \boldsymbol{x}_{N} \right)}\right] \right] \\
& = \mathbb{E}_{\boldsymbol{x}_{0}, \boldsymbol{x}_{1}, ..., \boldsymbol{x}_{N-1},\boldsymbol{x}_{N}}\left[- \log \frac{p_{\theta}\left( \boldsymbol{x}_{0}, \boldsymbol{x}_{1}, ..., \boldsymbol{x}_{N-1} \mid \boldsymbol{x}_{N} \right) }{q \left( \boldsymbol{x}_{1}, ..., \boldsymbol{x}_{N-1} \mid \boldsymbol{x}_{0}, \boldsymbol{x}_{N} \right)}\right] \\
& = \mathbb{E}_{q_{0:N}}\left[- \log \frac{p_{\theta}\left( \boldsymbol{x}_{0:N-1} \mid \boldsymbol{x}_{N} \right) }{q \left( \boldsymbol{x}_{1:N-1} \mid \boldsymbol{x}_{0}, \boldsymbol{x}_{N} \right)}\right].
\end{align*}    
\end{proof}

\subsection{The Proof of \cref{eq:property-1-assumption-1} in Sec. 3.2.1}

\begin{equation}
\begin{split}
    q \left( \boldsymbol{x}_{1:N-1} \mid \boldsymbol{x}_{0}, \boldsymbol{x}_{N} \right)
    = \prod_{t=1}^{N-1} q \left( \boldsymbol{x}_{t} \mid \boldsymbol{x}_{0}, \boldsymbol{x}_{t+1} \right).
\end{split}
\tag{15}
\end{equation}

\begin{proof}
According to Eq. (14), we have the following derivation:
\begin{align*}
& q \left( \boldsymbol{x}_{1:N-1} \mid \boldsymbol{x}_{0}, \boldsymbol{x}_{N} \right) \\
= & \prod_{t=1}^{N-1} q \left( \boldsymbol{x}_{t} \mid \boldsymbol{x}_{0}, \boldsymbol{x}_{t+1:N} \right) \\
= & \prod_{t=1}^{N-1} \frac{q \left( \boldsymbol{x}_{t}, \boldsymbol{x}_{t+2:N} \mid \boldsymbol{x}_{0}, \boldsymbol{x}_{t+1} \right)}{q \left( \boldsymbol{x}_{t+2:N} \mid \boldsymbol{x}_{0}, \boldsymbol{x}_{t+1} \right)} \\
= & \prod_{t=1}^{N-1} \frac{q \left(\boldsymbol{x}_{t+2:N} \mid \boldsymbol{x}_{0}, \boldsymbol{x}_{t+1}, \boldsymbol{x}_{t} \right)q \left( \boldsymbol{x}_{t} \mid \boldsymbol{x}_{0}, \boldsymbol{x}_{t+1} \right)}{q \left( \boldsymbol{x}_{t+2:N} \mid \boldsymbol{x}_{0}, \boldsymbol{x}_{t+1} \right)} \\
= & \prod_{t=1}^{N-1} \frac{q \left(\boldsymbol{x}_{t+2:N} \mid \boldsymbol{x}_{t+1} \right)q \left( \boldsymbol{x}_{t} \mid \boldsymbol{x}_{0}, \boldsymbol{x}_{t+1} \right)}{q \left( \boldsymbol{x}_{t+2:N} \mid \boldsymbol{x}_{t+1} \right)} \\
= & \prod_{t=1}^{N-1} q \left( \boldsymbol{x}_{t} \mid \boldsymbol{x}_{0}, \boldsymbol{x}_{t+1} \right).
\end{align*}
Equation (14) is applied in the fourth equation.
    
\end{proof}

\subsection{The Proof of \cref{eq:property-2-assumption-1} in Sec. 3.2.1}

\begin{equation}
\begin{split}
    q \left( \boldsymbol{x}_{t} \mid \boldsymbol{x}_{t+1:N}\right)
    = q \left( \boldsymbol{x}_{t} \mid \boldsymbol{x}_{t+1} \right).
\end{split}
\tag{16}
\end{equation}

\begin{proof}
\begin{align*}
& q \left( \boldsymbol{x}_{t} \mid \boldsymbol{x}_{t+1:N}\right) \\
=& \frac{q \left( \boldsymbol{x}_{t}, \boldsymbol{x}_{t+2:N} \mid \boldsymbol{x}_{t+1} \right)}{q \left(\boldsymbol{x}_{t+2:N} \mid \boldsymbol{x}_{t+1} \right)} \\
=& \frac{q \left( \boldsymbol{x}_{t} \mid \boldsymbol{x}_{t+1} \right) q \left(\boldsymbol{x}_{t+2:N} \mid \boldsymbol{x}_{t+1}, \boldsymbol{x}_{t} \right) }{q \left(\boldsymbol{x}_{t+2:N} \mid \boldsymbol{x}_{t+1} \right)}  \\
=& \frac{q \left( \boldsymbol{x}_{t} \mid \boldsymbol{x}_{t+1} \right) q \left(\boldsymbol{x}_{t+2:N} \mid \boldsymbol{x}_{t+1} \right) }{q \left(\boldsymbol{x}_{t+2:N} \mid \boldsymbol{x}_{t+1} \right)}  \\
=& q \left( \boldsymbol{x}_{t} \mid \boldsymbol{x}_{t+1} \right).
\end{align*}
Equation (14) is applied in the third equation.    
\end{proof}

\subsection{The Proof of \cref{eq:gaussian-q-posterior} and \cref{eq:gaussian-q-posterior-param} in Sec. 3.2.1}
\label{sec:proof-gaussian-posterior}

\begin{equation}
\begin{split}
    q \left( \boldsymbol{x}_{t} \mid \boldsymbol{x}_{0}, \boldsymbol{x}_{t+1} \right) \sim \mathcal{N} \left( \tilde{\boldsymbol{\mu}}_{t}, \tilde{\sigma}_{t} \boldsymbol{I}  \right), t = 0, 1, ..., N-1,
\end{split}
\tag{20}
\end{equation}
where 
\begin{equation}
\begin{split}
    \tilde{\boldsymbol{\mu}}_{t} &= \frac{\sigma_{t}^{2}}{\sigma_{t+1}^{2}} \boldsymbol{x}_{t+1} + \frac{\sigma_{t+1}^{2} - \sigma_{t}^{2}}{\sigma_{t+1}^{2}} \boldsymbol{x}_{0}, \\
    \tilde{\sigma}_{t}&= \frac{\sigma_{t}}{\sigma_{t+1}} \sqrt{\sigma_{t+1}^{2} - \sigma_{t}^{2}}.
\end{split}
\tag{21}
\end{equation}

\begin{proof}
Firstly, we have the following derivation according to Bayesian Equation.
\begin{align*}
 q\left( \boldsymbol{x}_{t} \mid \boldsymbol{x}_{0}, \boldsymbol{x}_{t+1} \right) &= \frac{q \left( \boldsymbol{x}_{t}, \boldsymbol{x}_{t+1} \mid \boldsymbol{x}_{0} \right)}{q \left( \boldsymbol{x}_{t+1} \mid \boldsymbol{x}_{0}\right)} \\
&= \frac{q \left( \boldsymbol{x}_{t} \mid \boldsymbol{x}_{0} \right) q \left( \boldsymbol{x}_{t+1} \mid \boldsymbol{x}_{t} \right) }{q \left( \boldsymbol{x}_{t+1} \mid \boldsymbol{x}_{0} \right)}.  
\end{align*}

Since $q \left( \boldsymbol{x}_{t} \mid \boldsymbol{x}_{0} \right)$, $q \left( \boldsymbol{x}_{t+1} \mid \boldsymbol{x}_{t} \right)$ and $q \left( \boldsymbol{x}_{t+1} \mid \boldsymbol{x}_{0} \right)$ all follow known Gaussian distribution, we can derive the analytical form for $q\left( \boldsymbol{x}_{t} \mid \boldsymbol{x}_{0}, \boldsymbol{x}_{t+1} \right)$. It is easy to verify that it also follows Gaussian distribution and the parameters are derived as \cref{eq:gaussian-q-posterior-param}.
\end{proof}

\subsection{The Proof of \cref{eq:gamma-xt} in Sec. 3.2.2}

\begin{equation}
\begin{split}
\boldsymbol{x}_{t} = \boldsymbol{\eta}_{t} \boldsymbol{x}_{0}, \quad \eta_{t, i} \sim \frac{1}{\alpha_{t}}  \mathcal{G} \left( \alpha_{t}, 1 \right),
\end{split}
\tag{31}
\end{equation}
where $t = 0, 1, ..., N-1$.

\begin{proof}
Firstly, we have a known conclusion from probability theory: Suppose $X \sim \mathcal{G} \left( \alpha, 1 \right), Y \sim \mathcal{G} \left( \beta, 1 \right)$, then 
\begin{equation*}
    \frac{X}{X + Y} \sim \mathcal{B} \left( \alpha, \beta \right),
\end{equation*}
and $X + Y$ is independent to $\frac{X}{X+Y}$.

Based on it, we have a corollary: Suppose $U$ is independent to $V$ and $U \sim \mathcal{B}( \alpha, \beta), V \sim \mathcal{G}(\alpha + \beta, 1)$, then
\begin{equation*}
    UV \sim \mathcal{G}(\alpha, 1).
\end{equation*}

Now, we prove \cref{eq:gamma-xt} according to the definition in Eq. (29) and Eq. (30). It is easy to verify that $\mathbf{x}_{1} \mid \mathbf{x}_{0}$ is satisfied. Assume $\mathbf{x}_{t-1} \mid \boldsymbol{x}_{0}$ follows $\boldsymbol{\eta}_{t-1} \boldsymbol{x}_{0}$, then
\begin{equation*}
\mathbf{x}_{t} \mid \boldsymbol{x}_{0} \sim \frac{\alpha_{t-1}}{\alpha_{t}} \boldsymbol{\zeta}_{t} \boldsymbol{\eta}_{t-1} \boldsymbol{x}_{0},
\end{equation*}
where $\boldsymbol{\eta}_{t-1} \sim \frac{1}{\alpha_{t-1}}  \mathcal{G} \left( \alpha_{t-1}, 1 \right)$, $\boldsymbol{\zeta}_{t} \sim \mathcal{B} \left( \alpha_{t+1}, \alpha_{t}- \right.$ $\left. \alpha_{t+1} \right)$. Since $\boldsymbol{\zeta}_{t}$ and $\boldsymbol{\eta}_{t-1}$ are independent, $\boldsymbol{\zeta}_{t} \boldsymbol{\eta}_{t-1} \sim \frac{1}{\alpha_{t-1}} \mathcal{G}(\alpha_{t}, 1)$. Considering the coefficient, \cref{eq:gamma-xt} is proved.
\end{proof}

\subsection{The Proof of \cref{eq:gamma-q-posterior-eq1} in Sec. 3.2.2}

\begin{equation}
\begin{split}
    \left(\frac{\alpha_{t} \boldsymbol{x}_{t} - \alpha_{t+1} \boldsymbol{x}_{t+1}}{\boldsymbol{x}_{0}}\right)_{i} \sim \mathcal{G} \left(\alpha_{t} - \alpha_{t+1}, 1 \right), t = 1, ..., N-1.
\end{split}
\tag{32}
\end{equation}

\begin{proof}
Similar to \cref{sec:proof-gaussian-posterior}, we still have
\begin{align*}
 q\left( \boldsymbol{x}_{t} \mid \boldsymbol{x}_{0}, \boldsymbol{x}_{t+1} \right) &= \frac{q \left( \boldsymbol{x}_{t}, \boldsymbol{x}_{t+1} \mid \boldsymbol{x}_{0} \right)}{q \left( \boldsymbol{x}_{t+1} \mid \boldsymbol{x}_{0}\right)} \\
&= \frac{q \left( \boldsymbol{x}_{t} \mid \boldsymbol{x}_{0} \right) q \left( \boldsymbol{x}_{t+1} \mid \boldsymbol{x}_{t} \right) }{q \left( \boldsymbol{x}_{t+1} \mid \boldsymbol{x}_{0} \right)}.  
\end{align*}
Since each component is independent, we prove \cref{eq:gamma-q-posterior-eq1} by components. We have known the following probability density functions:
\begin{align*}
    q \left( x_{t} \mid x_{0} \right) &= \frac{1}{\Gamma\left( \alpha_{t} \right)} \left(\frac{\alpha_{t} x_{t}}{x_{0}}\right)^{\alpha_{t} - 1} \exp \left\{- \frac{\alpha_{t} x_{t}}{x_{0}}\right\} \cdot \frac{\alpha_{t}}{x_{0} }, \\
    q \left( x_{t+1} \mid x_{0} \right) &= \frac{1}{\Gamma \left( \alpha_{t+1} \right)} \left(\frac{\alpha_{t+1} x_{t+1}}{x_{0}}\right)^{\alpha_{t+1} - 1} \\
    & \quad \quad \quad \cdot \exp \left\{- \frac{\alpha_{t+1} x_{t+1}}{x_{0}} \right\} \cdot \frac{\alpha_{t+1}}{x_{0}}, \\
    q \left( {x}_{t+1} \mid {x}_{t} \right) &= \frac{\Gamma\left( \alpha_{t} \right)}{\Gamma\left( \alpha_{t+1} \right)\Gamma\left( \alpha_{t} - \alpha_{t+1} \right)}  \left(\frac{\alpha_{t+1} x_{t}}{\alpha_{t} x_{t}} \right)^{\alpha_{t+1} - 1} \\
    & \quad \quad \quad \cdot \left( 1 - \frac{\alpha_{t+1} x_{t+1}}{\alpha_{t} x_{t}}  \right)^{ \alpha_{t} - \alpha_{t+1} - 1} \cdot \frac{\alpha_{t+1}}{\alpha_{t} x_{t}}.
\end{align*}
Applying the equations above, we have that
\begin{align*}
& \frac{q \left( \mathbf{x}_{t} \mid \mathbf{x}_{0} \right) q \left( \mathbf{x}_{t+1} \mid \mathbf{x}_{t} \right) }{q \left( \mathbf{x}_{t+1} \mid \mathbf{x}_{0} \right)} \\
= & \frac{1}{\Gamma\left( \alpha_{t} \right)} \left(\frac{\alpha_{t} x_{t}}{x_{0}}\right)^{\alpha_{t} - 1} \exp \left\{- \frac{\alpha_{t} x_{t}}{x_{0}} \right\} \cdot \frac{\alpha_{t}}{x_{0}} \\
& \quad \cdot \frac{\Gamma\left( \alpha_{t} \right)}{\Gamma\left( \alpha_{t+1} \right)\Gamma\left( \alpha_{t} - \alpha_{t+1} \right)} \left(\frac{\alpha_{t+1} x_{t+1}}{\alpha_{t} x_{t}} \right)^{\alpha_{t+1} - 1} \\
& \quad \cdot \left( 1 - \frac{\alpha_{t+1} x_{t+1}}{\alpha_{t} x_{t}}  \right)^{ \alpha_{t} - \alpha_{t+1} - 1} \cdot \frac{\alpha_{t+1}}{\alpha_{t} x_{t}} \\
& \quad / \left( \frac{\left(\frac{\alpha_{t+1} x_{t+1}}{x_{0}}\right)^{\alpha_{t+1} - 1}}{\Gamma\left( \alpha_{t+1} \right)}  \exp \left\{- \frac{\alpha_{t+1} x_{t+1}}{x_{0}}\right\} \cdot \frac{\alpha_{t+1}}{x_{0}} \right) \\
= & \frac{1}{\Gamma\left( \alpha_{t} - \alpha_{t+1} \right)} \frac{\left(\alpha_{t} x_{t} \right)^{\alpha_{t} - 1} \left(\alpha_{t+1} x_{t+1} \right)^{\alpha_{t+1} - 1} x_{0}^{\alpha_{t+1} - 1} }{ x_{0}^{\alpha_{t} - 1} \left(\alpha_{t} x_{t} \right)^{\alpha_{t+1} - 1} \left(\alpha_{t+1} x_{t+1} \right)^{\alpha_{t+1} - 1}} \\
& \quad \cdot \exp \left\{- \frac{\alpha_{t} x_{t}}{x_{0}} + \frac{\alpha_{t+1} x_{t+1}}{x_{0}}\right\}  \\
& \quad \cdot \left( 1 - \frac{\alpha_{t+1} x_{t+1}}{\alpha_{t} x_{t}}  \right)^{ \alpha_{t} - \alpha_{t+1} - 1} \frac{\alpha_{t}}{x_{0}} \cdot \frac{\alpha_{t+1}}{\alpha_{t} x_{t}} \cdot \frac{x_{0}}{\alpha_{t+1}} \\
= & \frac{1}{\Gamma\left( \alpha_{t} - \alpha_{t+1} \right)} \frac{\left(\alpha_{t} x_{t} \right)^{\alpha_{t} - \alpha_{t+1}} }{ x_{0}^{\alpha_{t} - \alpha_{t+1}}} \exp \left\{- \frac{\alpha_{t} x_{t} - \alpha_{t+1} x_{t+1}}{x_{0}} \right\} \\
& \quad \cdot \left( \frac{\alpha_{t} x_{t} - \alpha_{t+1} x_{t+1}}{\alpha_{t} x_{t}}  \right)^{ \alpha_{t} - \alpha_{t+1} - 1} \cdot \frac{1}{x_{t}} \\
= & \frac{\left(\frac{1}{x_{0}}\right)^{\alpha_{t} - \alpha_{t+1}}}{\Gamma\left( \alpha_{t} - \alpha_{t+1} \right)} \exp \left\{- \frac{\alpha_{t} x_{t} - \alpha_{t+1} x_{t+1}}{x_{0}}\right\} \\
& \quad \cdot \left( \alpha_{t} x_{t} - \alpha_{t+1} x_{t+1} \right)^{ \alpha_{t} - \alpha_{t+1} - 1} \cdot \alpha_{t}.
\end{align*}
Therefore, let $\tau_{t} = \frac{\alpha_{t} x_{t} - \alpha_{t+1} x_{t+1}}{x_0}$, then $\tau_{t}$ follows $\mathcal{G} \left(\alpha_{t} - \alpha_{t+1}, 1 \right)$. Thus, \cref{eq:gamma-q-posterior-eq1} is proved.
\end{proof}

\subsection{The Derivation of \cref{eq:gamma-kl} in Sec 3.2.2}

\begin{equation}
\begin{split}
&D_{\mathrm{KL}} \left( q \left( \boldsymbol{x}_{t} \mid \boldsymbol{x}_{0}, \boldsymbol{x}_{t+1} \right)  \|  p_{\theta} \left( \boldsymbol{x}_{t} \mid \boldsymbol{x}_{t+1} \right)\right) \\
=& \sum_{i} \left(\alpha_{t} - \alpha_{t+1} \right)\left( \log \frac{f_{\theta, i}}{{x}_{0,i}} + \frac{{x}_{0, i}}{f_{\theta, i}} - 1  \right).
\end{split}
\tag{35}
\end{equation}
Here, $f_{\theta}$ is the abbreviation of $f\left(\boldsymbol{x}_{t+1}, t+1; \theta \right)$. 
\begin{proof}
\begin{align*}
& D_{\mathrm{KL}} \left( q \left( \mathbf{x}_{t} \mid \mathbf{x}_{0}, \mathbf{x}_{t+1} \right)  \|  p_{\theta} \left( \mathbf{x}_{t} \mid \mathbf{x}_{t+1} \right)\right) \\
=& \int_{q} \log \frac{q \left( \mathbf{x}_{t} \mid \mathbf{x}_{0}, \mathbf{x}_{t+1} \right)}{p_{\theta} \left( \mathbf{x}_{t} \mid \mathbf{x}_{t+1} \right)} \mathrm{d} \mathbf{x}_{t} \\
=& \sum_{i} \int_{q} \left( \left(\alpha_{t} - \alpha_{t+1} \right)\log \frac{f_{\theta, i}}{x_{0,i}} - \left(\alpha_{t} x_{t} - \alpha_{t+1} x_{t+1} \right) \right. \\
& \quad \left.  \cdot \left(\frac{1}{x_{0,i}} - \frac{1}{f_{\theta, i}} \right) \right) \mathrm{d} \mathbf{x}_{t} \\
=& \sum_{i} \left(\alpha_{t} - \alpha_{t+1} \right)\log \frac{f_{\theta, i}}{x_{0,i}} - \mathbb{E} \left[x_{0,i} \tau_{t, i} \right] \left(\frac{1}{x_{0,i}} - \frac{1}{f_{\theta, i}} \right) \\
=& \sum_{i} \left(\alpha_{t} - \alpha_{t+1} \right)\log \frac{f_{\theta, i}}{x_{0,i}} - \left(\alpha_{t} - \alpha_{t+1}\right)x_{0,i} \left(\frac{1}{x_{0,i}} - \frac{1}{f_{\theta, i}} \right) \\
=& \sum_{i} \left(\alpha_{t} - \alpha_{t+1} \right)\left( \log \frac{f_{\theta, i}}{x_{0,i}} + \frac{x_{0,i}}{f_{\theta, i}} - 1  \right).
\end{align*}
    
\end{proof}

\subsection{The Proof of \cref{eq:gamma-mini} in Sec 3.2.2}

Suppose $f_{\theta^*}$ is the optimal function minimizing \cref{eq:gamma-kl}, it is also the optimal function for the following optimization problem:
\begin{equation}
\label{eq:gamma-mini}
\min_{f_{\theta}} \mathbb{E}_{q} \left\|f(\boldsymbol{x}_{t+1}, t+1; \theta) - \boldsymbol{x}_{0} \right\|_2^2. \tag{36}
\end{equation}

\begin{proof}
Minimizing KL divergence is equivalent to 
\begin{equation*}
\mathbb{E}_{q} \sum_i \left[ \left(\alpha_{i-1} - \alpha_{i} \right)\left( \log \frac{f(\boldsymbol{x}_{t+1}, t; \theta)_i}{x_{0}} + \frac{x_{0, i}}{f(\boldsymbol{x}_{t+1}, t; \theta)_i}  \right) \right].
\end{equation*}

Given $t$ and $\boldsymbol{x}_{t}$, according to \cite{xie2022trained} the optimal $f(\boldsymbol{x}_{t+1}, t; \theta)$ satisfies that
\begin{align*}
\arg \min_{\beta} \int q(\boldsymbol{x}_{0} \mid \boldsymbol{x}_{t+1}) \sum_i \left( \log \frac{\beta_i}{{x}_{0, i}} + \frac{{x}_{0, i}}{\beta_i}  \right) \mathrm{d} \boldsymbol{x}_{0} .
\end{align*}
We compute the gradient of $\beta$ for the right part and derive that
\begin{equation*}
\int q(\boldsymbol{x}_{0} \mid \boldsymbol{x}_{t+1}) \left( \frac{1}{\beta} - \frac{\boldsymbol{x}_{0}}{\beta ^{2}}  \right) \mathrm{d} \boldsymbol{x}_{0}.    
\end{equation*}
Let the gradient to be $\boldsymbol{0}$ (we can neglect the situation where $\beta = 0$), we have
\begin{equation*}
\int q(\boldsymbol{x}_{0} \mid \boldsymbol{x}_{t+1})  \left(\beta - \boldsymbol{x}_{0}  \right) \mathrm{d} \boldsymbol{x}_{0}  = \boldsymbol{0}.
\end{equation*}
When $\beta = \mathbb{E} \left[ \boldsymbol{x}_{0} \mid \boldsymbol{x}_{t+1} \right]$, it is the optimal solution. Therefore, the optimal $f(\mathbf{x}_{t+1}, t; \theta)$ should be $\mathbb{E} \left[ \boldsymbol{x}_{0} \mid \boldsymbol{x}_{t+1} \right]$. That is to say, training $f(\mathbf{x}_{t+1}, t; \theta)$ by minimizing KL divergence is equivalent to minimize 
\begin{equation*}
\mathbb{E}_{q} \left\|f(\boldsymbol{x}_{t+1}, t; \theta) - \boldsymbol{x}_{0} \right\|_2^2.
\end{equation*}

\end{proof}

\subsection{The Proof of \cref{eq:poisson-xt} in Sec 3.2.3}

\begin{equation}
\begin{split}
\boldsymbol{x}_{t} \mid \boldsymbol{x}_{0} \sim \frac{\mathcal{P}  \left( \lambda_{t} \boldsymbol{x}_{0} \right)}{\lambda_{t}},
\end{split}
\tag{41}
\end{equation}
where $t = 1, ..., N-1$.

\begin{proof}
Firstly, \cref{eq:poisson-xt} holds when $t = N$. Assume that $\mathbf{x}_{t+1} \mid \boldsymbol{x}_{0}$ holds, then
\begin{align*}
\mathbf{x}_{t} \mid \boldsymbol{x}_{0} & \sim \frac{\lambda_{t+1} \mathbf{x}_{t+1} \mid \boldsymbol{x}_{0} + \mathcal{P} \left( \left(\lambda_{t} - \lambda_{t+1}\right) \boldsymbol{x}_{0} \right)}{\lambda_{t}} \\
&\sim \frac{\mathcal{P} \left( \left(\lambda_{t+1}\right) \boldsymbol{x}_{0} \right) + \mathcal{P} \left( \left(\lambda_{t} - \lambda_{t+1}\right) \boldsymbol{x}_{0} \right)}{\lambda_{t}} \\
&\sim \frac{\mathcal{P}  \left( \lambda_{t} \boldsymbol{x}_{0} \right)}{\lambda_{t}} .
\end{align*}
   
\end{proof}

\subsection{The Derivation of \cref{eq:poisson-kl} in Sec 3.2.3}

\begin{equation}
\begin{split}
&D_{\mathrm{KL}} \left( q \left( \boldsymbol{x}_{t} \mid \boldsymbol{x}_{0}, \boldsymbol{x}_{t+1} \right)  \|  p_{\theta} \left( \boldsymbol{x}_{t} \mid \boldsymbol{x}_{t+1} \right)\right) \\
=& \sum_i \left( \log {x}_{0,i} - \log f_{\theta, i} \right) \left( \lambda_{t} - \lambda_{t+1} \right) {x}_{0, i} \\
& \quad \quad - \left( \lambda_{t} - \lambda_{t+1} \right) \left({x}_{0, i} - f_{\theta, i} \right),
\end{split}
\tag{46}
\end{equation}
where $f(\boldsymbol{x}_{t+1}, \boldsymbol{x}_{N}, t+1; \theta)$ is denoted as $f_{\theta}$ for simplicity.

\begin{proof}
Let $\tau_{t} = \lambda_{t} \mathbf{x}_{t} - \lambda_{t+1} \mathbf{x}_{t+1}$. For simplicity, we regard all vectors as variables. Then, we have the following derivation
\begin{align*}
& D_{\mathrm{KL}} \left( q \left( {x}_{t} \mid {x}_{0}, {x}_{t+1} \right)  \|  p_{\theta} \left( {x}_{t} \mid {x}_{t+1} \right)\right) \\
=& \sum_{{x}_{t}} q \left( {x}_{t} \mid {x}_{0}, {x}_{t+1}\right) \log \frac{q \left( {x}_{t} \mid {x}_{0}, {x}_{t+1} \right)}{p_{\theta} \left( {x}_{t} \mid {x}_{t+1} \right)} \\
=& \sum_{\tau_{t}} \frac{\left( \lambda_{t} - \lambda_{t+1}\right)^{\tau_{t}} {x}_{0}^{\tau_{t}}} {\tau_{t} !} e^{- \left( \lambda_{t} - \lambda_{t+1} \right){x}_{0} }  \\
& \quad \cdot \log \dfrac{\dfrac{\left( \lambda_{t} - \lambda_{t+1}\right)^{\tau_{t}} {x}_{0}^{\tau_{t}}} {\tau_{t} !} e^{- \left( \lambda_{t} - \lambda_{t+1} \right){x}_{0} }}{\dfrac{\left( \lambda_{t} - \lambda_{t+1}\right)^{\tau_{t}} f_{\theta}^{\tau_{t}}} {\tau_{t} !} e^{- \left( \lambda_{t} - \lambda_{t+1} \right)f_{\theta}}} \\
=& \sum_{\tau_{t}} \frac{\left( \lambda_{t} - \lambda_{t+1}\right)^{\tau_{t}} {x}_{0}^{\tau_{t}}} {\tau_{t} !} e^{- \left( \lambda_{t} - \lambda_{t+1} \right){x}_{0} } \\
& \quad \cdot \left( \left( \log {x}_{0} - \log f_{\theta} \right) \tau_{t} - \left( \lambda_{t} - \lambda_{t+1} \right) \left({x}_{0} - f_{\theta} \right) \right) \\
=& \left( \log {x}_{0} - \log f_{\theta} \right) \mathbb{E} \left[ \tau_{t} \right] - \left( \lambda_{t} - \lambda_{t+1} \right) \left({x}_{0} - f_{\theta} \right) \\
=& \left( \log {x}_{0} - \log f_{\theta} \right) \left( \lambda_{t} - \lambda_{t+1} \right) {x}_{0} - \left( \lambda_{t} - \lambda_{t+1} \right) \left({x}_{0} - f_{\theta} \right).
\end{align*}
Therefore, \cref{eq:poisson-kl} is proved.

\end{proof}

\subsection{The Proof of \cref{eq:poisson-mini} in Sec 3.2.3}

Suppose $f_{\theta^*}$ is the optimal function minimizing \cref{eq:poisson-kl}, it is also the optimal function for the following optimization problem:
\begin{equation}
\begin{split}
\min_{f_{\theta}} \mathbb{E}_{q} \left\|f(\boldsymbol{x}_{t+1}, \boldsymbol{x}_{N}, t+1; \theta) - \boldsymbol{x}_{0} \right\|_2^2.
\end{split}
\tag{47}
\end{equation}

\begin{proof}
Minimizing KL divergence is equivalent to 
\begin{equation*}
\begin{split}
\mathbb{E}_{\boldsymbol{x}_{0}, \boldsymbol{x}_{t+1}} & \sum_i \left( \log {x}_{0,i} - \log f_{\theta, i} \right) \left( \lambda_{t} - \lambda_{t+1} \right) {x}_{0, i} \\
& \quad \quad - \left( \lambda_{t} - \lambda_{t+1} \right) \left({x}_{0, i} - f_{\theta, i} \right).
\end{split}
\end{equation*}

Given $t$ and $\boldsymbol{x}_{t}$, according to \cite{xie2022trained} the optimal $f(\boldsymbol{x}_{t+1}, \boldsymbol{x}_{N}, t; \theta)$ satisfies that
\begin{align*}
\arg & \min_{\boldsymbol{s}} \int q(\boldsymbol{x}_{0} \mid \boldsymbol{x}_{t+1}, \boldsymbol{x}_{N})\sum_i  \left( \left( \log {x}_{0, i} - \log {s}_i \right) \right. \\
& \quad \cdot \left. \left( \lambda_{t} - \lambda_{t+1} \right) {x}_{0, i} - \left( \lambda_{t} - \lambda_{t+1} \right) \left({x}_{0, i} - {s}_i \right) \right) \mathrm{d} \boldsymbol{x}_{0}. 
\end{align*}
We compute the gradient of $\boldsymbol{s}$ for the right part and derive that
\begin{equation*}
\int q(\boldsymbol{x}_{0} \mid \boldsymbol{x}_{t+1}, \boldsymbol{x}_{N}) \left( -\left( \lambda_{t} - \lambda_{t+1} \right) \frac{\boldsymbol{x}_{0}}{\boldsymbol{s}} + \left( \lambda_{t} - \lambda_{t+1} \right) \right) \mathrm{d} \boldsymbol{x}_{0}.  
\end{equation*}
Let the gradient to be $\boldsymbol{0}$, we have
\begin{equation*}
\begin{split}
& \int q(\boldsymbol{x}_{0} \mid \boldsymbol{x}_{t+1}, \boldsymbol{x}_{N}) \left( -\frac{\boldsymbol{x}_{0}}{\boldsymbol{s}} + 1 \right) \mathrm{d} \boldsymbol{x}_{0} \\
= & -\frac{\mathbb{E}\left[ \boldsymbol{x}_{0} \mid \boldsymbol{x}_{t+1}, \boldsymbol{x}_{N} \right] }{\boldsymbol{s}} + 1 = \boldsymbol{0}.
\end{split}
\end{equation*}
When $\boldsymbol{s} = \mathbb{E} \left[ \boldsymbol{x}_{0} \mid \boldsymbol{x}_{t+1}, \boldsymbol{x}_{N} \right]$, it is the optimal solution. Therefore, the optimal $f(\mathbf{x}_{t+1},\boldsymbol{x}_{N}, t; \theta)$ should be $\mathbb{E} \left[ \boldsymbol{x}_{0} \mid \boldsymbol{x}_{t+1}, \boldsymbol{x}_{N}, \right]$. That is to say, training $f(\mathbf{x}_{t+1}, \boldsymbol{x}_{N}, t; \theta)$ by minimizing KL divergence is equivalent to minimize 
\begin{equation*}
\mathbb{E}_{q} \left\|f(\boldsymbol{x}_{t+1}, \boldsymbol{x}_{N}, t; \theta) - \boldsymbol{x}_{0} \right\|_2^2.
\end{equation*}

\end{proof}

\section{Experimental Details}
\label{sec:experiment_detail}

\paragraph{Parameters Sequence}

For simplicity, suppose $\left\{\alpha_{t}\right\}$ is the parameters sequence used for constructing the diffusion process. Let $\sigma_{\alpha_t}$ represent the standard deviation of $\mathbf{x}_{t} \mid \boldsymbol{x}_{0}$ from $t = 0$ to $t = N$ where $N$ is the diffusion steps. We have that $\sigma_{\alpha_0} = 0$ and $\sigma_{\alpha_{N}}$ is known. We set 
\begin{equation*}
    \sigma_{\alpha_{t}} = \frac{t}{N} * \sigma_{\alpha_{N}}
\end{equation*}
and ascertain the value of $\alpha_t$ according to $\sigma_{\alpha_{t}}$.

\begin{figure*}
  \centering
  \includegraphics[width=\linewidth]{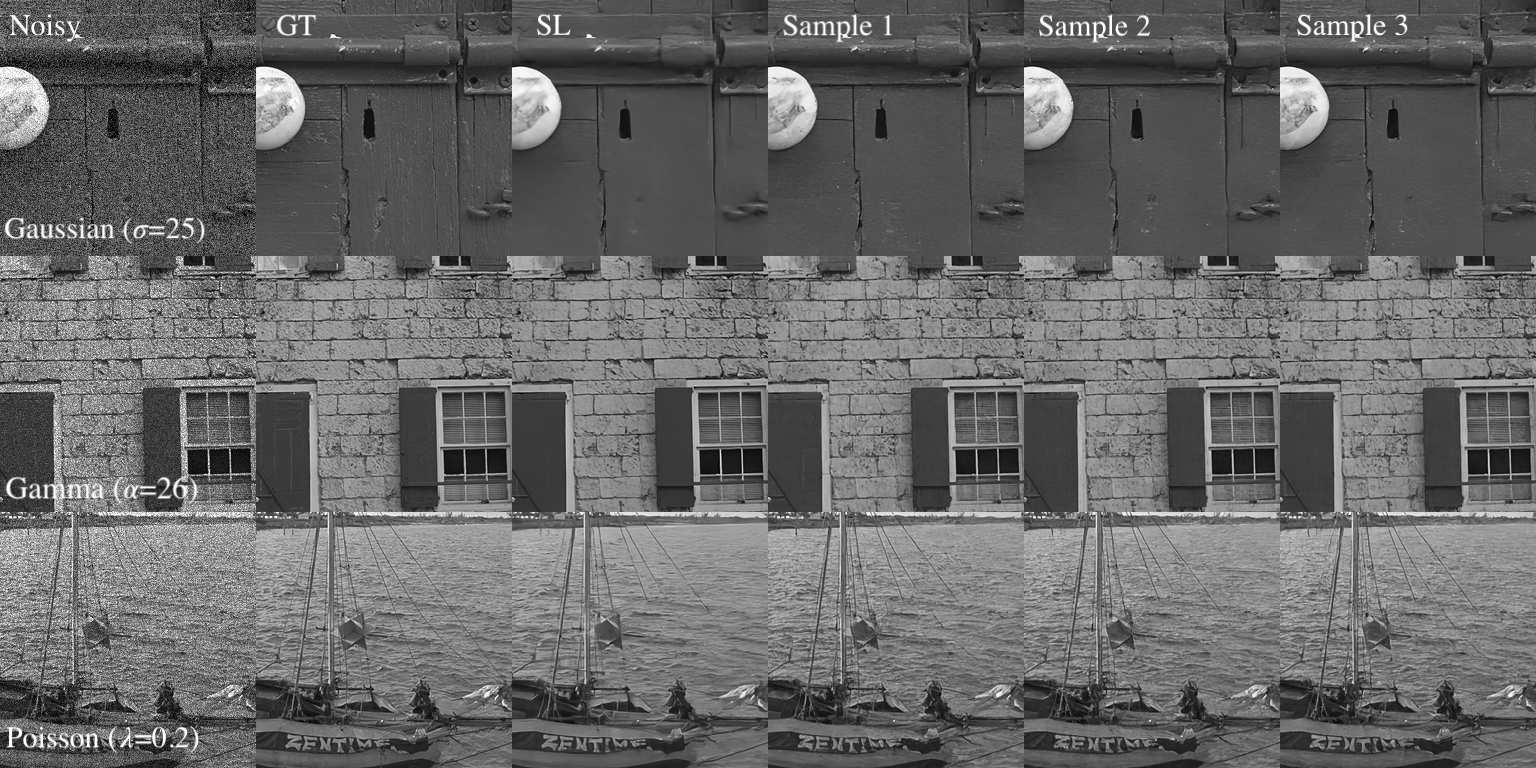}
  \caption{Qualitative comparison using Kodak dataset. From top to bottom is: Gaussian noise with $\sigma=25$, Gamma noise with $\alpha=26$, and Poisson noise with $\lambda=0.2$. Noisy: noisy image, GT: ground-truth, SL: supervised learning. The last three columns are generated different samples using $N = 10$.}
  \label{fig:results-N-10}
\end{figure*}

\paragraph{The Effect of Steps $N$}
For Gaussian noise with $\sigma=25$, Gamma noise with $\alpha = 26$ and Poisson noise with $\lambda = 0.2$, we also conduct experiments where $N = 10$ and compare the result to $N = 20$. The result is shown in \cref{tab:result-metric-N-10} and \cref{fig:results-N-10}. From the metrics and visual quality, our method also achieve satisfied performance even with $N = 10$.

\begin{table*}
  \centering
  \begin{tabular}{@{}l|ccc|ccc@{}}
    \toprule
    Noise Model & \multicolumn{3}{c}{Kodak}\vline & \multicolumn{3}{c}{CSet9} \\
    \hline
     & SL & Samples & Mean of Samples & SL & Samples & Mean of Samples \\
     \hline
    Gaussian, $\sigma = 25$
       & 30.53 / 0.816 & 28.62 / 0.761 &  30.48 / 0.821 & 29.41 / 0.843 & 27.61 / 0.800 &  29.44 / 0.849 \\

    Gamma, $\alpha = 26$
       & 31.18 / 0.849 & 29.35 / 0.796 &  31.16 / 0.851 & 29.68 / 0.854 & 27.98 / 0.807 &  29.76 / 0.856 \\
       
    Poisson, $\lambda = 0.2$
       & 30.95 / 0.834 & 28.88 / 0.776 &  30.46 / 0.835 & 29.68 / 0.851 & 27.69 / 0.800 &  29.25 / 0.853 \\

    \bottomrule
  \end{tabular}
  \caption{Quantitative comparison in terms of PNSR (dB) / SSIM for different noise model on Kodak and CSet9 datasets. The number of diffusion steps is set as $10$. We compare supervised learning (SL) to our method. "Samples" means the generated results while "Mean of Samples" means the average of $100$ samples for each noisy images. }
  \label{tab:result-metric-N-10}
  
\end{table*}

\end{document}